%% file: Main.tex
\documentclass[letterpaper, 10 pt, conference]{ieeeconf}
\IEEEoverridecommandlockouts                   

\usepackage{graphicx}   
\graphicspath{ {images/} }
\usepackage{lineno,hyperref}
\usepackage{comment}
\usepackage{amsmath}
\usepackage{algorithm}
\usepackage{algorithmic}
\usepackage{amssymb}
\usepackage{hyperref}
\usepackage{comment}
\usepackage{accents}
\usepackage{braket, relsize, nccmath} 
\usepackage{subcaption}
\usepackage{microtype}
\usepackage{xcolor}

\usepackage{caption}

\newcommand{\tildeu}{\tilde{u}}

\newcommand{\dzero}{\tilde{d}}

\newtheorem{Prop}{Proposition}
\newtheorem{thm}{Theorem}

\newtheorem{rmk}{Remark}

\newtheorem{definition}{Definition}

\title{\LARGE \bf A Coupled Control Barrier Function Approach to Deadlock-Free Multi-Robot Conflict Resolution}
\usepackage{times}
\usepackage{mathtools}

\author{
Gary Rostomyan \qquad
Ketan Savla \qquad
Petros A. Ioannou
\thanks{The authors are with the University of Southern California. Email:\texttt{\{rostomya,ksavla,ioannou\}@usc.edu}. This work was supported in part by METRANS PSR-24-55. K. Savla has a financial interest in Xtelligent, Inc.}}

\begin{document}
\maketitle
\input{abstract} 
\input{Intro}
\input{PRELIMINARIES}
\input{COUPLED_CBF}

\input{MULTI-AGENT_COLLISION_AVOIDANCE}

\input{Simulation}
\input{Conclusion}
% \input{Appendix}

\bibliographystyle{IEEEtran}
\bibliography{l4dc2026-references}

\end{document}

%% file: abstract.tex
\begin{abstract}
Many robotic applications involve scenarios where multiple robots interact and the decisions of one robot influence the behavior of others. This paper presents a quadratic programming (QP)-based feedback controller for multi-robot conflict resolution with input constraints. Safety is enforced using coupled Control Barrier Function (CBF), while a \textit{yielding matrix} encodes how robots influence each other. Unlike existing methods with fixed or hard priorities, the proposed framework enables yielding preferences while guaranteeing deadlock-free operation by design. Sufficient conditions are established for the controller to be locally Lipschitz.
The optimal CBF sign-indefinite slack is shown to admit a finite uniform bound, which is compensated through an enlarged CBF safety margin to obtain a forward invariant set contained within the desired safe set. The analysis also identifies an irreducible slack component due to finite maximum deceleration that cannot be eliminated by increasing the slack penalty. Finally, sufficient conditions on the yielding matrix and nominal controller are derived to exclude deadlock states. 
Simulations demonstrate adaptive coordination across yielding preferences and robustness to robot failures.
\end{abstract}
% \begin{keywords}%
% control barrier functions, decentralized control, game theory, multiagent systems.
% \end{keywords}

%% file: Intro.tex
\section{Introduction}
\begin{figure}
    \centering
    \includegraphics[width=2.5in]{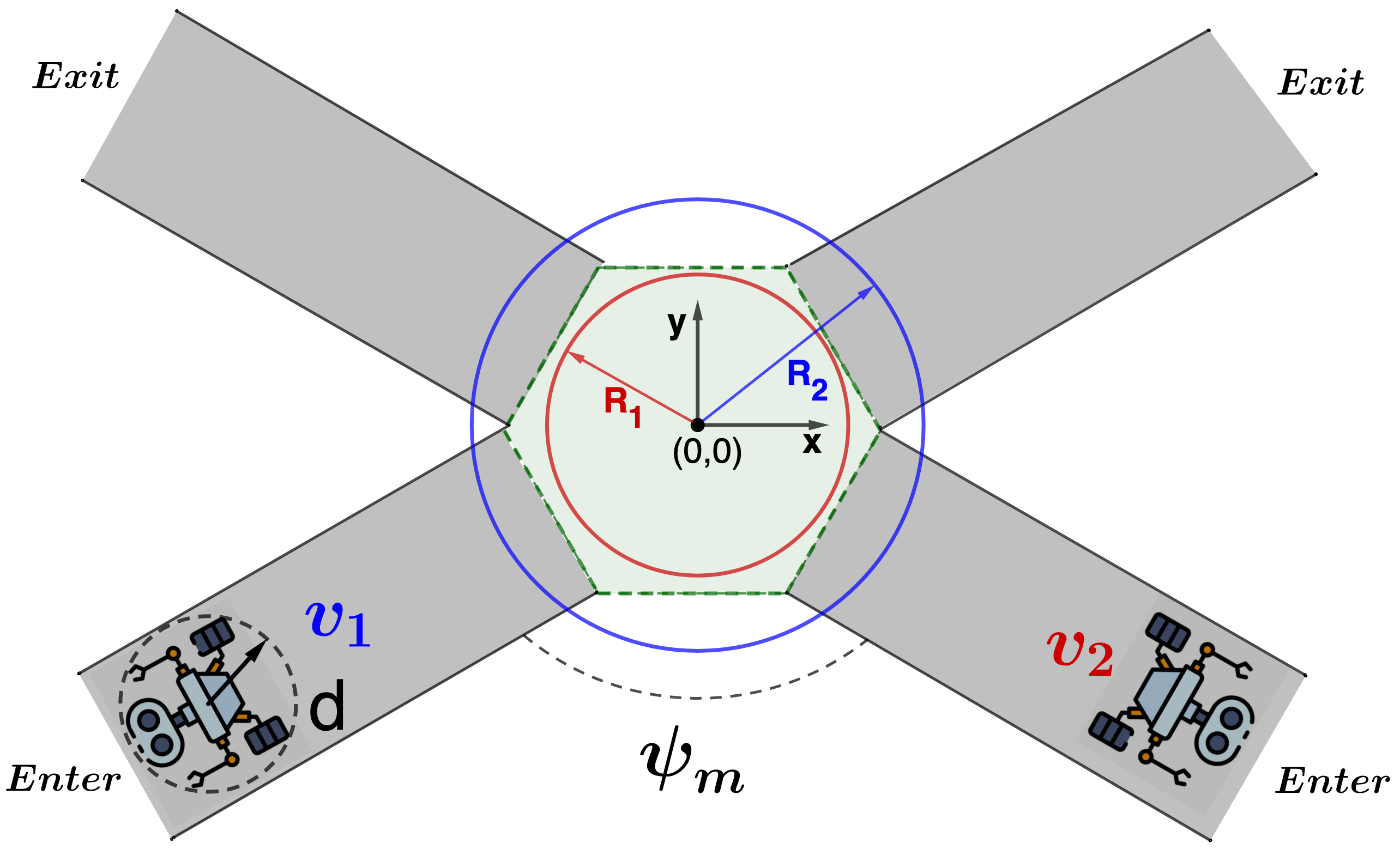}
    \caption{Two mobile robots, modeled as disks of radius $d$, navigate narrow corridors, through a conflict zone (green shaded hexagon), where $\psi_m$ denotes the minimum pairwise intersection angle among the corridors. 
    The unsafe region is a disk with a dynamically changing radius
    $\min\{R_1,R_2\}$, where each $R_i$ equals the robot’s stopping distance based on its velocity $v_i$, plus a desired safety margin. No two robots may occupy an unsafe region simultaneously.}
    \label{Intro_collision}
     \vspace{-0.5mm}
\end{figure}

% As modern logistics and manufacturing operations continue to expand, achieving high efficiency and accuracy in storing and retrieving goods within warehouses has become increasingly critical. Mobile robots play a key role in boosting warehouse performance by automating material transportation, order picking, and replenishment activities \cite{deKoster2023}. By autonomously transferring items between storage zones, picking stations, and shipping docks, mobile robots reduce the need for manual labor and streamline warehouse operations. Their adaptability, scalability, and flexibility allow warehouses to significantly improve productivity, accuracy, and overall customer satisfaction \cite{Grover-Ashraf}.
% Consequently, many modern warehousing operations employ fleets of mobile robot systems. In such systems, each mobile robot performs assigned tasks—often with varying priorities—that generally involve picking up a load at a specified origin and delivering it to a designated destination.
As logistics and warehouse operations continue to expand, mobile robots have become essential for automating material transport, order picking, and replenishment, thereby improving productivity, accuracy, and operational efficiency \cite{deKoster2023, Grover-Ashraf}. Modern warehouses deploy mobile robot fleets for prioritized transport tasks. When robots enter shared conflict zones along planned corridor routes, collisions and deadlocks may occur (Fig.~\ref{Intro_collision}). Thus, safety and deadlock-free
operation are key design objectives. 
% In addition to avoiding collisions among multiple mobile robots, autonomous warehouse operations may also require coordinated robot-to-robot picking. This concept involves collaborative item transfer between robots. 
% A representative scenario illustrated in Fig.~\ref{Intro_collision} includes a picker mobile robot R3 equipped with a manipulator arm and a staging mobile robot R6 that transports items to predefined rendezvous zone.
% Many robotic applications involve interactions between
% multiple agents. For instance, two quadcopters may need to
% interact and implicitly coordinate to successfully navigate in a
% shared three-dimensional space (Fig. 1). An autonomous car
% may need to interact with human-driven cars or pedestrians to
% navigate an intersection. Planning in such interactive settings
% is in general challenging due to the feedback interactions
% between the agents. An agent’s state and action will affect the
% state and actions of the other agents, i.e., agents are coupled by
% their intentions and actions. In this
% \begin{figure}
%     \centering
%     \includegraphics[width= 0.8\linewidth]{fig/Intro_fig_coll_rendev.png}
%     \caption{An example of task assignments for mobile robots, where robots R1–R3 and R3–R6 must avoid collisions within the conflict zone indicated by the red dots, while robots R3 and R6 are tasked with performing a robot-to-robot picking task in the designated rendezvous zone indicated by the blue dot. 
%     }
%     \label{Intro_collision}
% \end{figure}
The conflict resolution problem shown in Fig.~\ref{Intro_collision} involves interactions among multiple autonomous robots. Planning in such interactive environments is inherently complex since each robot’s state and control affect the others.

Multi-robot interactions can be modeled as differential games \cite{Sadigh2016PlanningFA, Sadigh2016InformationGA}. 
However, computing equilibria requires solving coupled optimal control problems, making these approaches computationally intractable for real-time applications \cite{basar1999dynamic}.
Prior works have also modeled multi-robot interactions as dynamic potential games, enabling the computation of a generalized Nash equilibrium by solving a single optimal control problem rather than multiple coupled problems \cite{Bhatt-Negar}-\cite {Jia_Negar}. However, these approaches require restrictive cost structures (e.g., action-independent inter-agent costs) to preserve the potential game property, thus precluding general non-potential costs. Moreover, they compute open-loop Nash equilibria, even when implemented in a receding-horizon fashion, and provide no guarantees of recursive feasibility.

In contrast, the proposed approach supports non-potential cost functions and yields a well-defined, optimization-based feedback controller. Since state parametrized optimization-based controllers are not necessarily locally Lipschitz \cite{MESTRES2025101098}, establishing this property is essential for guaranteeing existence and uniqueness of the closed-loop smooth solution.
In \cite{ames2016control}, local Lipschitz continuity is established for a QP-based controller with two constraints by explicitly characterizing all possible active sets and deriving closed-form Lagrange multipliers. This approach, however, does not readily scale to QPs with many constraints, for which the number of possible active sets grows exponentially.
In contrast, we provide verifiable sufficient conditions for the local Lipschitz continuity of a QP-based controller under multiple input constraints. We further present the unexplored interplay among safety, deadlock-free operation, and well-posed closed-loop motion under multiple input constraints.

Earlier work in \cite{BORRMANN} proposed an optimization-based centralized controller for multi-robot collision avoidance; however, it neither accounts for coupling costs that encode yielding relationships between robots nor establishes feasibility over a set of initial conditions.
% In the decentralized approach, the authors adopt a restrictive assumption regarding agent interactions, assuming that each agent treats all others as maintaining constant velocities.
% Our proposed centralized controller enables \lq soft\rq\ prioritization among robots, where \lq soft\rq\ priority refers to priority encoded in the optimization cost rather than imposed as a hard ordering constraint, while also guaranteeing conflict resolution under control input constraints.
The proposed feedback controller enables the specification of pairwise yielding preferences among robots without imposing a fixed a priori crossing order through the conflict zone. These yielding preferences are encoded in the optimization cost rather than enforced through hard ordering constraints, thereby influencing the robots’ motions toward the desired yielding behavior.
Unlike the ad hoc braking strategy in \cite{wang2017safety}, which may induce deadlocks when safety constraints become infeasible, and reactive deadlock-resolution methods, our framework balances local Lipschitz continuity of the QP-based controller and forward invariance of a subset of the safe set with deadlock prevention through a conservative design margin.
% Recent work \cite{Jankovic_2022} proposes a centralized collision avoidance algorithm but doesn't account for input constraints or a yielding mechanism encoded by \lq soft\rq\ priorities.
Recent work \cite{Jankovic_2022} analyzes two-robot deadlock resolution using a closed-form QP solution without input constraints. It uses fixed pairwise separation for safety constraints, whereas our formulation uses speed-dependent stopping-distance margins.
Moreover, \cite{Jankovic_2022} identifies a deadlock as an unstable saddle equilibrium, whereas our analysis provides constructive conditions on the controller parameters that eliminate deadlock for $N$ robots.

% This paper presents a centralized QP-based multi-robot control framework with \lq soft\rq\ priority assignments. 
% Note, the \lq soft\rq\ priorities specify pairwise yielding behavior rather than a fixed a priori order for crossing the conflict zone, and influence the robots’ motion only when yielding is needed to avoid an imminent collision.
The main contributions are threefold. First, sufficient conditions are established  
for a locally Lipschitz feedback controller with input constraints that guarantees safety and deadlock-free operation by design, with yielding preferences encoded in a \textit{yielding matrix}.
% Second, sufficient conditions are established for forward invariance of the safe set under control input constraints, and for ensuring that the resulting feedback controller is locally Lipschitz continuous, thereby guaranteeing the existence and uniqueness of the closed-loop solution.
Second, to maintain the desired safety in the presence of sign-indefinite CBF slack, a compensation mechanism is introduced by enlarging the  safety margin to guarantee forward invariance of a set contained within the desired safe set. Finally, simulations demonstrate that multi-robot  coordination adapts to assigned yielding preferences, resulting in yielding behavior, while also maintaining coordination robustness against robot failures.

%% file: PRELIMINARIES.tex
\section{PRELIMINARIES}\label{PRELIMINARIES}
% \vspace{-2mm}
\subsection{Notation}
% In this paper, the set of non-negative real numbers is denoted by $\mathbb{R}_+$. A continuous function $\alpha:[0,b) \rightarrow \mathbb{R}_+$ is called a class-$\mathcal{K}$ function when it is continuous, strictly increasing, and $\alpha(0) = 0$, 
A continuous function $\alpha: (-b, c) \rightarrow \mathbb{R}$ is said to belong to extended class $\mathcal{K}$ for some $b, c>0$ if it is strictly increasing and $\alpha(0) = 0$ (see \cite[Definition 2]{ames2016control}). 
In this paper, the linear extended class $\mathcal{K}$ function $\alpha(s) = \gamma s$ is used, where $\gamma >0$ is the gain parameter.
The Lie derivative of the function $h:\mathbb{R}^n \rightarrow \mathbb{R}$ with respect to the vector field $f: \mathbb{R}^n \rightarrow \mathbb{R}^n$ is defined as $\mathcal{L}_f h = \frac{\partial h(\mathbf{x})}{\partial \mathbf{x}} f(\mathbf{x})$. 
% A function $f: \mathcal{D} \rightarrow \mathbb{R}^n$ is locally Lipschitz if for any compact set $\mathcal{D}_0 \subseteq \mathcal{D}$ there is a constant $L$ such that $\|f(x) - f(y)\|_2 \leq L \| \mathbf{x} - \mathbf{y}\|_2$ for all $\mathbf{x}, \mathbf{y} \in \mathcal{D}_0$.
The complement of a set $S$ is denoted as $\bar{S}$. Let $[N]=\{1, \dots, N\}$ denote the index set of the $N$ robots. The operator $\operatorname{col}(\cdot)$ denotes vertical concatenation of its arguments, and $[x]_+:= \max\{x,0\}$.

\subsection{Background}
The proposed approach builds on the CBF framework introduced in \cite{wieland2007constructive} and later extended to optimization-based safety control in \cite{ames2016control}. 
Consider the control-affine system:
\begin{equation}\label{affinesystemclosed}
    \dot{\mathbf{x}}= f(\mathbf{x})+g(\mathbf{x})\mathbf{u},
\end{equation}
where $f\!:\mathbb{R}^n \to \mathbb{R}^n$ and $g\!:\mathbb{R}^n \to \mathbb{R}^m$ are continuous and locally Lipschitz.
% \begin{equation}\label{affinesystem}
%     \dot{\mathbf{x}}= f(\mathbf{x})+g(\mathbf{x})\mathbf{u},
% \end{equation}
% The state is denoted by $\mathbf{x} \in \mathcal{D} \subset \mathbb{R}^n $, and the control input $\mathbf{u} \in \mathcal{U} \subset \mathbb{R}^m$ is selected from the set of admissible controls. 
% The goal of CBF is to design a controller $\mathbf{u}=k(\mathbf{x})$, such that the closed-loop system: 
% \begin{equation}\label{affinesystemclosed}
%     \dot{\mathbf{x}}= f(\mathbf{x})+g(\mathbf{x})\mathbf{u},
% \end{equation}
% meets specific safety requirements.
If the feedback controller $\mathbf{u} = k(\mathbf{x})$ is continuous and locally Lipschitz, then the closed-loop system \eqref{affinesystemclosed} has a unique solution $\mathbf{x}(t)$ on maximum interval of existence $I(\mathbf{x}_0)= (a,b)$ for initial condition $\mathbf{x}_0$. 
% Since state parametrized optimization-based feedback controllers are not necessarily locally Lipschitz \cite{robinson2009generalized}, it is important to establish this property to guarantee the existence and uniqueness of solutions of the closed-loop system \eqref{affinesystemclosed}. 
    % When  the maximal interval extends to $b=+\infty$, $f_{cl}$ is called forward complete.
\begin{definition}
The set $\mathcal{C}$ is forward invariant under the closed-loop system \eqref{affinesystemclosed} with $\mathbf{u} = k(\mathbf{x})$ if every trajectory starting in $\mathcal{C}$ remains in $\mathcal{C}$ for all $t \in I(\mathbf{x}_0)$.
The system \eqref{affinesystemclosed} is deemed safe with respect to the set $\mathcal{C}$ if $\mathcal{C}$ is forward invariant under the closed-loop system \eqref{affinesystemclosed} with $\mathbf{u} = k(\mathbf{x})$. 
\end{definition}

Let the safe set be $\mathcal{C}: = \{\mathbf{x} \in \mathcal{D} \ | \ h(\mathbf{x}) \geq 0 \},$ where $h : \mathcal{D} \subset \mathbb{R}^n \rightarrow \mathbb{R}$ is continuously differentiable function.
A key result \cite[Corollary 2]{ames2016control} establishes that if a locally Lipschitz controller $\mathbf{u}= k(\mathbf{x}) \in \mathcal{U}$, where $\mathcal{U}$ denotes the set of admissible control inputs, satisfies
$\mathcal{L}_f h(\mathbf{x}) + \mathcal{L}_gh(\mathbf{x})k(\mathbf{x})  + \gamma h(\mathbf{x})\geq0$
for all $\mathbf{x} \in \mathcal{C}$, then $\mathcal{C}$ is forward invariant under the resulting closed-loop system \eqref{affinesystemclosed}.

%% file: COUPLED_CBF.tex
\section{PROBLEM FORMULATION}\label{Sec_veh_model} 

\textit{Kinematic model}:
Mobile robots are represented in the X–Y plane as non-holonomic agents constrained to move along straight corridors (Fig.~\ref{Intro_collision}), maintaining a constant steering angle. Each robot’s state is defined as
$\boldsymbol{\xi} = [x, y, v]^T$, where $x$ and $y$ specify the its position along the longitudinal and lateral axes of a fixed Cartesian frame with its origin at the center of a conflict area, as shown in Fig.~\ref{Intro_collision}, which are assumed to remain bounded, and $v$ denotes its speed. As such, the kinematics of each mobile robot are modeled in control-affine form as:
% \begin{equation}\label{dynamics1}
% \dot{\boldsymbol{\xi}} = 
% \underbrace{\begin{bmatrix}
% v\cos(\theta)\\
% v\sin(\theta) \\
% 0 \\
% \end{bmatrix}}_{f(\boldsymbol{\xi})}
% +
% \underbrace{\begin{bmatrix}
% 0 \\
% 0  \\
% 1 
% \end{bmatrix}}_{g(\boldsymbol{\xi})}
% \tildeu,
% \end{equation}
\begin{equation}\label{dynamics1}
\dot{\boldsymbol{\xi}}
=
f(\boldsymbol{\xi})+g(\boldsymbol{\xi})\tilde{u},
\end{equation}
% \begin{equation} \label{dynamics1}
% \begin{cases}
% \dot{x} = v  cos(\theta),\\
% \dot{y} = v sin(\theta), \\
% % \dot{\theta} = \frac{v}{L} \tan(\psi), \\
% \dot{v} = u, 
% \end{cases}
% \end{equation}
where $f(\boldsymbol{\xi})= [v\cos\phi, \  v\sin\phi, \ 0]^\top$, $g(\boldsymbol{\xi})= [0, \
0,\ 1 ]^\top,$ and 
$\tildeu \in [u_m,u_M]$ is the constrained acceleration input applied to the robot. Here, $u_m<0$ and $u_M>0$ denote the maximum deceleration and acceleration, respectively, 
and $\phi$ is the constant heading angle describing the robot’s body-fixed forward axis orientation relative to the $x$-axis of the Cartesian reference frame. 
For example, in Fig.~\ref{Intro_collision}, $\phi_1 = 45^{\circ}$ and $\phi_2 = 135^{\circ}$ remain constant during the robots’ motion. For simplicity, all mobile robots in this work are assumed to be identical  and follow the model in \eqref{dynamics1}. 
% Additionally, all robots are constrained to the same range of admissible control inputs, $u \in [u_m, u_M]$.

To ensure $v\geq0$, a low-speed correction is introduced. Let $v_{\mathrm{th}}>0$ and $\tau>0$ be a design parameters, chosen such that $\tau\geq-\frac{u_m}{v_{\mathrm{th}}}$.
The feedback controller $u(\boldsymbol{\xi})$, which is designed to ensure safety and deadlock-free operation while maintaining control performance, is then post-processed as
$\tildeu=\max\{u(\boldsymbol{\xi}),-\tau v\}$. 
Therefore, for $v \ge v_{th}$, $-\tau v \leq u_m \leq u(\boldsymbol{\xi})$, and thus $\tildeu = u(\boldsymbol{\xi})$. For $0 \leq v \leq v_{th}$, $\tildeu \geq - \tau v$, and therefore $\dot{v} \geq -\tau v$, which guarantees $v(t) \geq 0$ whenever $v(0) \geq 0$. Hence, $-\tau v(t)\leq 0<u_M$, and if $u(\boldsymbol{\xi})\in[u_m,u_M]$, it follows that $\tildeu \in [u_m,u_M]$. 
% the post-processed rule is inactive for $v\geq v_{\mathrm{th}}$, so that $\tildeu=u(\boldsymbol{\xi})$, while at $v=0$ it guarantees $\tildeu \geq0$, thereby preventing the speed from becoming negative. 
% To ensure $v\geq 0$, the feedback controller $u(\boldsymbol{\xi})$, that is constrained to the same range of admissible control inputs, $u \in [u_m, u_M]$, is post-processed according to $u_{\mathrm{ap}}=\max\{u(\boldsymbol{\xi}),-\tau v\},
% $ where $\tau$ is chosen such that $-\tau v\leq u_m$ for all $v\geq v_{\mathrm{th}}$, with threshold $v_{\mathrm{th}}\ll1$. Consequently, the post-clipping rule is inactive for $v\geq v_{\mathrm{th}}$, while at $v=0$ it guarantees $u_{\mathrm{ap}}\geq0$, thereby preventing the speed from becoming negative.
Moreover, $\tilde{u}$ remains locally Lipschitz since the maximum of two locally Lipschitz functions is locally Lipschitz. The aggregated state and input vectors of the $N$ mobile robots are denoted as $\boldsymbol{\xi}_a=col(\boldsymbol{\xi}_i)_{i=1}^N \in\mathbb{R}^{3N}$ and $\tilde{\mathbf{u}}=\max\{\mathbf{u}_a(\boldsymbol{\xi}_a),-\tau\mathbf{v}_a\}\in\mathbb{R}^N$, respectively, where $\mathbf{u}_a=col(u_i)_{i=1}^N$ is the aggregated feedback controller, $\mathbf{v}_a=col(v_i)_{i=1}^N$ is the vector of aggregated speeds, and the $\max$ is taken componentwise. 
The aggregated $N$ robot model is
\begin{equation}\label{stacked_system}
    \dot{\boldsymbol{\xi}}_a = f_a(\boldsymbol{\xi}_a) + g_a(\boldsymbol{\xi}_a)\mathbf{\tildeu},
\end{equation}
where $f_a(\boldsymbol{\xi}_a) = \operatorname{col} \big (f(\boldsymbol{\xi}_i) \big )_{i=1}^N \in \mathbb{R}^{3N}$, and
$g_a(\boldsymbol{\xi}_a) = \operatorname{diag}\big(g(\boldsymbol{\xi}_1), \dots, g(\boldsymbol{\xi}_N) \big) \in \mathbb{R}^{3N\times N}$ is a block-diagonal matrix whose diagonal blocks are given by $g(\boldsymbol{\xi}_i), \ i \in [N]$. 
Since each $f(\boldsymbol{\xi}_i)$ and $g(\boldsymbol{\xi}_i), i \in [N]$, is locally Lipschitz for all $i\in [N]$, so are $f_a(\boldsymbol{\xi}_a)$ and $g_a(\boldsymbol{\xi}_a)$. We next define the safe set $S_{sf}$ for $N$ mobile robots.

\textit{Safety specifications}:
For a desired minimum safety-margin parameter $\dzero>0$, define the time-varying conflict region $\mathcal{B}(t;\dzero):= \{p \in \mathbb{R}^2 \; | \; \|p\|_2< R_{m}(t;\dzero) \}$, $R_m(t;\dzero)= \min_{i \in [N]} R_i(t;\dzero)$, where $R_i(t;\dzero):= v_i^2/(2|u_b|)+\dzero$, as shown in Fig.~\ref{Intro_collision}. Note, $R_i(t; \dzero)$ is the sum of robot $i$’s stopping distance and $\dzero$. The configuration is deemed safe with respect to $\dzero$ if, at every time $t$, at most one robot center lies in $\mathcal{B}(t;\dzero)$. The desired minimum safety margin is required to satisfy $\dzero > \frac{d}{\sin (\psi_m/2)}>0$, where $d$ denotes the robot radius and 
$\psi_m$ the minimum pairwise intersection angle among the corridors.
This condition ensures that the center-to-center distance between any pair of stopped robots along the corridors exceeds twice the robot radius, $2d$.
Moreover, the deceleration $u_b$ used in the safety specification is assumed to satisfy $u_b \in [u_m, 0)$, resulting in a conservative estimate of the stopping distance thereby providing an additional safety margin.
The unsafe configuration parametrized by  $\dzero$ is defined as $S_{\mathrm{usf}}(\dzero) = \bigcup\limits_{ i<j} S_i(\dzero) \cap S_j(\dzero)$, where
% $
%     S_{i} = \big\{(x_i,y_i, v_i) \ | \ (x_i - x_p)^2 + (y_i- y_p)^2 - \frac{v_i^4}{4u_m^2} < 0 \big\}.
% $
$S_i(\dzero): = \{ \boldsymbol{\xi}_i \; | \; \boldsymbol{\xi}_i^\top P \boldsymbol{\xi}_i - \big( \tfrac{(\mathbf{e}_3^\top \boldsymbol{\xi}_i)^2}{2|u_b|} +\dzero \big)^2< 0 \}$,
with $P = \mathrm{diag}(1,1,0)$ and $\mathbf{e}_3 =[ 0, 0, 1]^\top$.
The parameterized safe set is
\begin{equation}\label{complex_safety}
S_{sf}(\dzero) = \bar{S}_{usf}(\dzero) = \bigcap\limits_{i<j} \big (\bar{S}_{i}(\dzero)\cup \bar{S}_{j}(\dzero) \big).
\end{equation} 
It is important to note that the safety condition in \eqref{complex_safety} is formulated as a composition of multiple sets.
The objective is to represent all such compositions as the 0-superlevel set of a single candidate CBF, which is achieved by applying the composition methods in \cite{Molnar_Ames}. Notice $\bar{S}_{i}(\dzero) \cup \bar{S}_{j}(\dzero) = \{ (\boldsymbol{\xi}_i, \boldsymbol{\xi}_j) \in \mathbb{R}^{6} \ | \ \max\{h(\boldsymbol{\xi}_i), h(\boldsymbol{\xi}_j)\} \geq 0 \}$ can be expressed as the 0-superlevel set of the function $\max\{h(\boldsymbol{\xi}_i), h(\boldsymbol{\xi}_j)\}$, where the set $\bar{S}_{i}(\dzero)$ is the 0-superlevel set of function
$h(\boldsymbol{\xi}_i; \dzero) = \boldsymbol{\xi}_i^\top P \boldsymbol{\xi}_i - \big (\tfrac{(\boldsymbol{e}_3^\top \boldsymbol{\xi}_i)^2}{2|u_b|} + \dzero \big )^2$, $i \in [N]$. 
Although the union of sets can be represented by a maximum function, the latter is not continuously differentiable and thus unsuitable as a candidate CBF. We therefore use the smooth under-approximation
$h_{ij}(\boldsymbol{\xi}_a;\kappa, \dzero)= \frac{1}{\kappa} \ln \big (e^{\kappa h(\boldsymbol{\xi}_i; \dzero)} + e^{\kappa h(\boldsymbol{\xi}_j; \dzero)} \big) - \frac{\ln(2)}{\kappa}$,
where $h_{ij}(\boldsymbol{\xi}_a;\kappa, \dzero)\leq \max\{h(\boldsymbol{\xi}_i; \dzero), h(\boldsymbol{\xi}_j; \dzero)\}$ and $\kappa >0$ is the smoothing factor \cite{lindemann2018control}.
The 0-superlevel set of $h_{ij}(\boldsymbol{\xi}_a; \kappa, \dzero)$ satisfies $\{ \boldsymbol{\xi}_a \ | \
h_{ij}(\boldsymbol{\xi}_a; \kappa, \dzero) \geq 0 \} \subseteq \bar{S}_{i}(\dzero) \cup \bar{S}_{j}(\dzero)$ 
and converges to 
$\bar{S}_{i}(\dzero) \cup \bar{S}_{j}(\dzero)$ as $\kappa \to \infty$, since $
\lim_{\kappa \to \infty} h_{ij}(\boldsymbol{\xi}_a;\kappa, \dzero)
=\max\left\{h(\boldsymbol{\xi}_i; \dzero),h(\boldsymbol{\xi}_j; \dzero)\right\}
$.
Combining all $\binom{N}{2}$ pairwise $h_{ij}$ using a smooth under-approximation of the minimum, yields the aggregate candidate CBF $ h_a(\boldsymbol{\xi}_a;\kappa, \dzero)= -\frac{1}{\kappa} \ln \big ( \sum\limits_{i < j} e^{-\kappa h_{ij}(\boldsymbol{\xi}_a;\kappa, \dzero)} \big)  \leq \min\limits_{i < j} h_{ij}(\boldsymbol{\xi}_a;\kappa, \dzero)$. By construction, $\{\boldsymbol{\xi}_a \ | \ h_{a}(\boldsymbol{\xi}_a; \kappa, \dzero) \geq 0 \}  \subseteq\mathcal S_{\mathrm{sf}}(\dzero)$.
% Also, note that the intersection of sets $\bigcap\limits_{1 \le i < j \le N}\mathcal{P}_{ij} \subseteq \bigcap\limits_{1 \le i < j \le N} \big (\neg S_{i} \cup \neg S_{j} \big)$, which, in turn, can be represented as a 0-superlevel set of 
% $\min\limits_{1 \le i < j \le N} h_{max}(\boldsymbol{\xi}_i, \boldsymbol{\xi}_j)$, where the minimum is computed over $\binom{N}{2}$ number of pairs $(i,j)$. 
% Similarly, since the minimum function is not continuously differentiable, we propose a candidate $h_a(\boldsymbol{\xi}_a): \mathbb{R}^{3N} \to \mathbb{R}$ that provides a smooth under-approximation of the minimum function as 
% \begin{equation*}    
%   h_a(\boldsymbol{\xi}_a)= -\frac{1}{\kappa} \ln \bigg ( \sum\limits_{1 \le i < j \le N} e^{-\kappa h_{max}(\boldsymbol{\xi}_i, \boldsymbol{\xi}_j)} \bigg)  \leq \min\limits_{1 \le i < j \le N} h_{max}(\boldsymbol{\xi}_i, \boldsymbol{\xi}_j).
% \end{equation*}
Next, we derive the Lie derivatives of $h_a$ along the vector fields $f_a$ and $g_a$. Clearly, $\frac{\partial h_a}{\partial \boldsymbol{\xi}_i} = \beta_{i} \frac{\partial h(\boldsymbol{\xi}_i)}{\partial \boldsymbol{\xi}_i}$, where $\beta_i = \sum_{j \neq i} \mu_{ij} \frac{e^{-\kappa h_{ij}(\boldsymbol{\xi}_a)}}{\sum_{i < j} e^{-\kappa h_{ij}(\boldsymbol{\xi}_a)}}$, and $\mu_{ij} = \frac{e^{\kappa h(\boldsymbol{\xi}_i)}}{e^{\kappa h(\boldsymbol{\xi}_i)} + e^{\kappa h(\boldsymbol{\xi}_j)}}$, $i \in [N]$.
Thus, $\mathcal{L}_{f_a} h_a = \sum\limits_{i=1}^N 2\beta_i v_i\big (x_i\cos\phi_i + y_i\sin\phi_i\big)$ and
% \begin{equation*}\label{Lie_along_f}   
%     \begin{split}
%         &\mathcal{L}_{f_a} h_a(\boldsymbol{\xi}_a) =\\
%         &=\sum\limits_{i=1}^2 \beta_i \big(2P\boldsymbol{\xi}_i  - \frac{2(\boldsymbol{e}_3^\top \boldsymbol{\xi}_i)}{|u_b|}\left(\frac{(\boldsymbol{e}_3^\top \boldsymbol{\xi}_i)^2}{2|u_b|} + d_0 \right)e_3 \big)^\top f(\boldsymbol{\xi}_i) = \\
%         &=\sum\limits_{i=1}^2 2\beta_i v_i\big (x_i\cos(\theta_i) + y_i\sin(\theta_i)\big).
%         % \frac{\partial h_a}{\partial \xi_i} = \frac{\sum\limits_{\substack{j=1 \\ j \neq i}}^N w_{ij} \alpha_{ij} \frac{\partial h(\xi_i)}{\partial \xi_i}}{\sum\limits_{1\leq i <j \leq N} w_{ij}}   
%     \end{split}
% \end{equation*}
 % \frac{2}{\sum_{i=1}^{2} \eta(\mathbf{\xi}_i)} \sum_{i=1}^{2} \eta(\mathbf{\xi}_i) v_i \big((x_i - x_p) \cos{\theta_i} + \\
 %        &+(y_i - y_p)\sin{\theta_i}\big)
% where the Jacobian is
% $\frac{\partial h_s (\xi_s)}{\partial \xi_s} = \frac{2}{\sum_{i=1}^2 e^{\kappa h(\xi_i)}} 
% \begin{pmatrix}
%     2e^{\kappa h(\xi_1)}(x_1 -x_p)\\
%     2e^{\kappa h(\xi_1)}(y_1 -y_p)\\
%     -e^{\kappa h(\xi_1)}\frac{v_1^3}{2u_m^2}\\
%     2e^{\kappa h(\xi_2)}(x_2 -x_p)\\
%     2e^{\kappa h(\xi_2)}(y_2 -y_p)\\
%     -e^{\kappa h(\xi_2)}\frac{v_2^3}{2u_m^2}\\
% \end{pmatrix}$.
$\mathcal{L}_{g_a} h_a  = \operatorname{col} \big ( - \frac{2\beta_i v_i}{|u_b|}(\frac{v_i^2}{2|u_b|} + \dzero) \big)_{i=1}^N$.
% \begin{equation}
% \begin{split}
% \mathcal{L}_{f_a} h_a &= \sum_{i=1}^N 2\beta_i v_i \big( x_i\cos(\theta_i) + y_i\sin(\theta_i) \big), \\
% \mathcal{L}_{g_a} h_a &= \left[ - \frac{2\beta_i v_i}{|u_b|} \left( \frac{v_i^2}{2|u_b|} + d_0 \right) \right]_{i=1}^N.
% \end{split}
% \end{equation}
    % \frac{\partial h_a}{\partial \xi_i} = \frac{\sum\limits_{\substack{j=1 \\ j \neq i}}^N w_{ij} \alpha_{ij} \frac{\partial h(\xi_i)}{\partial \xi_i}}{\sum\limits_{1\leq i <j \leq N} w_{ij}}    
% where $\frac{\partial h_a}{\partial 
% \boldsymbol{\xi}_i} g(\boldsymbol{\xi}_i) =
% % \beta_i \bigg(2P\boldsymbol{\xi}_i - \frac{2(e_3^\top \boldsymbol{\xi}_i)}{|u_m|}\left(\frac{(e_3^\top \boldsymbol{\xi}_i)^2}{2|u_m|} + d_0 \right)e_3 \bigg)^\top g(\boldsymbol{\xi}_i) = 
% - \frac{2\beta_i v_i}{|u_b|}\big(\frac{v_i^2}{2|u_b|} + d_0 \big),$ $i \in [N]$.
Since $\mathbf{v}_a \succeq 0$, then $\mathcal{L}_{g_a} h_a \preceq 0$, a property used in the following sections. 
% For the desired minimum safety margin $\tilde d$, specialize the candidate CBF by setting $d_s = \tilde{d}$. 
Define $\tilde  h(\boldsymbol{\xi}_a):= h_a(\boldsymbol{\xi}_a;\kappa,\tilde d)$ and its 0-superlevel set as
$\tilde {\mathcal{C}}:= \{\boldsymbol{\xi}_a \ | \ \tilde h(\boldsymbol{\xi}_a) \geq 0 \}$. 
% Let the 0-superlevel set of the continuously differentiable function $h_a(\boldsymbol{\xi}_a)$ be $\mathcal{C}:= \{\boldsymbol{\xi}_a \ | \  h_a(\boldsymbol{\xi}_a) \geq 0 \}$ and note that $\mathcal{C} \subseteq S_{sf}$. 

\textit{Deadlock-free operation}:
Another important consideration is ensuring deadlock-free operation of the $N$ robot system. Following \cite{wang2017safety}, the deadlock set is defined as follows.
\begin{definition}\label{Deadlock_free_def}
Multi-robot system \eqref{stacked_system} is in deadlock if $\mathcal{J}_D: = \{\boldsymbol{\xi}_a \;|\; \mathbf{v}_a = \mathbf{0}, \tilde{\mathbf{u}}(\boldsymbol{\xi}_a)=0,$ and no  robot has cleared the unsafe region\} $\neq \varnothing$.
% , where $s_i=x_i\cos\theta_i+y_i\sin\theta_i>0$ indicates that the robot \(i\) has passed the center of the unsafe region. 
% A multi-robot feedback controller \(\mathbf{u}_a(\boldsymbol{\xi}_a)\) is \emph{deadlock-free} if \(\mathcal{J}_D\) is not forward invariant under the closed-loop dynamics \eqref{stacked_system}. 
\end{definition}
Each robot has a nominal controller $u_{0i}(\boldsymbol{\xi}_i)$, $i \in [N]$, designed to achieve the desired performance (e.g., traversing a corridor at the maximum allowable speed),  but without  necessarily guaranteeing safety or deadlock-free operation. 

\textit{Problem Statement}:
Given the multi-robot system \eqref{stacked_system}, the safe set
$\tilde {\mathcal{C}}$, and the nominal controller $\mathbf{u}_0=col(u_{0i})_{i=1}^N$, we seek a locally Lipschitz feedback controller 
$\mathbf{u}_a(\boldsymbol{\xi}_a) \in [u_m,u_M]^N$ that minimizes a weighted combination of the deviation from $\mathbf{u}_0$ and a coupling cost encoding the desired pairwise yielding behavior, while (i) guaranteeing that  $\boldsymbol{\xi}_a(t) \in \tilde {\mathcal{C}}$ for all $t \geq 0$ and 
(ii) ensuring $\mathcal{J}_D =\varnothing$.

%% file: MULTI-AGENT_COLLISION_AVOIDANCE.tex
\section{PROPOSED CONTROLLER}\label{centeralized}
To achieve the control objectives stated in the problem statement, a centralized feedback controller is proposed that collects the robots’ states and computes
$\mathbf{u}_a(\boldsymbol{\xi}_a)$ as
the solution to a QP problem parameterized by $\boldsymbol{\xi}_a$. Safety is enforced 
through the barrier function  $h_0(\boldsymbol\xi_a):=h_a(\boldsymbol\xi_a;\kappa,d_0)$ implemented in the QP, constructed using a conservative  margin $d_0>\tilde d$, which balances two objectives:
 % The QP enforces safety through a coupled CBF constraint,
% $\mathcal{L}_{f_a}h_a(\boldsymbol{\xi}_a) + \mathcal{L}_{g_a}h_a(\boldsymbol{\xi}_a)\mathbf{u}_a \geq -\alpha(h_a(\boldsymbol{\xi}_a))$
% while balancing two objectives: 
minimizing the deviation from the nominal controller, measured by $J_1(\mathbf{u}_a):=\|\mathbf{u}_a -\mathbf{u}_0\|_2^2$, and minimizing the coupling cost $J_2(\mathbf{u}_a):= (\mathbf{u}_a- u_m \mathbf{1}_N)^\top Q (\mathbf{u}_a +u_m \mathbf{1}_N)$, where $Q=[q_{ij}] \in \mathbb{R}^{N \times N}$ is the \textit{yielding matrix} with $q_{ii}=0$. 
The shifts by $u_m$ ensure that the asymmetric part of $Q$ influences the optimizer. For $N=2$, the coupling cost is, up to a constant,
$(q_{12}+q_{21})u_1u_2 +
u_m(q_{12}-q_{21})(u_1-u_2).$
The symmetric first term couples the inputs without distinguishing which robot yields, while the asymmetric second term determines the yielding direction. 
Since $u_m<0$, if $q_{12}<q_{21}$, minimizing the asymmetric second  term biases the
solution toward $u_1<u_2$, causing robot~1 to apply less acceleration and thus yields. The sign of $q_{12}-q_{21}$ determines which robot yields.  
% Since \(u_m<0\), if $q_{12}<q_{21}$, minimizing the second term favors $u_1<u_2,$
% so robot 1 to brake more and thus yields more.
Accordingly, for a given state $\boldsymbol{\xi}_a$, the  controller solves the following QP problem to determine $\mathbf{u}_a(\boldsymbol{\xi}_a)$:
\begin{align}
&\min_{\mathbf{u}_a,\delta}
  \frac{\nu}{2}J_1(\mathbf{u}_a)
  +\frac{1-\nu}{2}J_2(\mathbf{u}_a) +\frac{p\delta^2}{2}=\notag\\
 &\qquad =\min_{\mathbf{u}_a,\delta}
  \frac12(\mathbf{u}_a^\top H\mathbf{u}_a+p\delta^2)
  +l(\boldsymbol{\xi}_a)^\top\mathbf{u}_a,
  \label{QP_center_1}\\
&\text{s.t. }\boldsymbol g(\mathbf{u}_a, \delta;\boldsymbol{\xi}_a)
 \geq\boldsymbol 0, \notag
\end{align}
% \begin{equation}\label{QP_center_1}
%     \begin{split}
%         &\min_{\mathbf{u}_a, \delta} \frac{\nu}{2} \|\mathbf{u}_a - \mathbf{u}_0\|_2^2 
%          + \frac{1-\nu}{2}(\mathbf{u}_a - u_m \mathbf{1}_2)^\top Q(\mathbf{u}_a + u_m\mathbf{1}_2) + \\
%          &+ \frac{p\delta^2}{2}=\min_{\mathbf{u}_a,\delta} \frac{1}{2}(\mathbf{u}^\top_a H \mathbf{u}_a + p\delta^2) + l^\top \mathbf{u}_a \\
%         &\text{s.t. } \boldsymbol{g}(\mathbf{u}_a,\boldsymbol{\xi}_a) \geq \boldsymbol{0},
%     \end{split}
% \end{equation}
where constraints $\boldsymbol{g}(\mathbf{u}_a,\delta; \boldsymbol{\xi}_a) = col (g^{cbf}, \boldsymbol{g}^{lb}, \boldsymbol{g}^{ub})$ are defined as: $g^{cbf} = \mathcal{L}_{f_a} h_0+ \mathcal{L}_{g_a} h_0^\top \mathbf{u}_a +\gamma h_0 + \delta$, $\boldsymbol{g}^{lb} = \mathbf{u}_a - u_m\mathbf{1}_N$, $\boldsymbol{g}^{ub} = u_M\mathbf{1}_N - \mathbf{u}_a$, and where, after expansion and term collection, the cost function is written in standard quadratic form with $H = \nu I+ (1-\nu)Q$, $l(\boldsymbol{\xi}_a) = -\nu \mathbf{u}_0(\boldsymbol{\xi}_a) + (1-\nu) u_m Q_{sk}\mathbf{1}_N$, $I \in \mathbb{R}^{N \times N}$ is the identity matrix, $Q_{sk} = \frac{(Q-Q^\top)}{2}$, and $\nu \in (0,1)$ is a trade-off parameter that balances the importance of following nominal control inputs $\mathbf{u}_0$ against the yielding preferences encoded by $Q$.
In the QP \eqref{QP_center_1}, $\delta \in \mathbb{R}$ is the slack variable, $p>0$ is the penalty coefficient, and $\mathbf{1}_N$ is the all-ones vector.
Note, the feedback controller $\mathbf{u}_a(\boldsymbol{\xi}_a)$ depends on the control parameters $(Q,\kappa,\gamma,p,u_b,d_0)$, whose selection is discussed later.
\begin{rmk}
   Unlike conventional CBF-based QPs with weighted tracking costs $\sum_i w_i (u_i-u_{0i})^2$, where $w_i>0$, the cost in \eqref{QP_center_1} employs a \textit{yielding matrix} $Q$ to encode pairwise yielding preferences which cannot be captured by the weighted tracking costs. Specifically, when the QP constraints are inactive, $\sum_i w_i (u_i-u_{0i})^2$ is minimized at $\mathbf{u}_a(\boldsymbol{\xi}_a)= \mathbf{u}_0$ regardless of the values of $w_i$, and therefore cannot induce yielding.
   In contrast, the proposed coupling cost contains asymmetric terms that shift the optimizer $\mathbf{u}_a(\boldsymbol{\xi}_a)$ from $\mathbf{u}_0$, promoting anticipatory coordination according to yielding preferences even before the QP constraints become active.
  
\end{rmk}
% \begin{rmk}
%     % relative to its nominal control, with the lower-priority robot absorbing more of the required deviation. This priority is not inherently associated with yielding; its effect depends on the respective nominal controls.
% \end{rmk}

\begin{rmk}
    Introducing a sign-indefinite slack $\delta$ into the CBF constraint $g^{\mathrm{cbf}}$ ensures the feasibility of the QP \eqref{QP_center_1} for any $\boldsymbol{\xi}_a$, and local Lipschitz continuity of the controller $\mathbf{u}_a(\boldsymbol{\xi}_a)$, as established in Theorem \ref{theorem2}; however, the forward invariance of the set $\{\boldsymbol{\xi}_a \ | \ h_{0}(\boldsymbol{\xi}_a) \geq 0 \}$ is no longer trivially guaranteed. Instead, the forward invariance of its enlarged set is guaranteed by Theorem \ref{forwardinv}.
\end{rmk}
Note that
$\mathbf{u}_a^\top Q \mathbf{u}_a = \mathbf{u}_a^\top Q_{sp}\mathbf{u}_a$, where $Q_{sp} =\frac{Q+Q^\top}{2}$. 
Therefore, the Hessian of the cost function with respect to $\mathbf{u}_a$ can be equivalently expressed as $H = \nu I+ (1-\nu)Q_{sp}$. 
The QP \eqref{QP_center_1} admits a unique solution if $H \succ 0$.
\begin{Prop}\label{prop2}
    $H \succ 0$ if any of the following holds:
    \begin{enumerate}
        \item $Q_{sp}$ is positive semidefinite, 
        \item the smallest eigenvalue of $Q_{sp}$ is $\lambda_{min}(Q_{sp}) > -\frac{\nu}{1-\nu}$,
        \item $\sum_{j \neq i} |\frac{q_{ij} + q_ {ji}}{2}| <\frac{\nu}{1-\nu}$, for all $i \in [N]$.
    \end{enumerate}
\end{Prop}
\textbf{Locally Lipschitz continuity:} 
Since the robots operate in a bounded spatial domain, it suffices to show that $v_i$ is bounded for all $i \in [N]$ 
to ensure that $\boldsymbol{\xi}_a$ remains in a compact set $\mathcal X$.
\begin{Prop}\label{compactness} 
If $H\succ 0$,  $q_{ij}\geq0$ for all $i \neq j$, and for each robot
$i\in[N]$, there exists a finite $V_i\geq0$ such that
$u_{0i}(\boldsymbol\xi_i)\le 
\frac{(1-\nu)}{\nu}u_m
\sum_{j\ne i}q_{ij},$
whenever $v_i\ge V_i$, then
$v_i(t)\leq\max\{v_i(0),V_i\},$ for all $i\in[N]$, $\forall t\geq0.$
\end{Prop}
\begin{proof}
Fix any robot $i\in[N]$. We show that the solution of 
\eqref{QP_center_1} satisfies
$u_i\leq 0
\ \text{whenever } v_i\geq V_i.$
Since $\delta\in\mathbb{R}$ and $u_m<u_M$, the QP \eqref{QP_center_1}  satisfies Slater's condition for every
$\boldsymbol{\xi}_a$. Since the QP is convex, the Karush–Kuhn–Tucker (KKT) conditions are
necessary and sufficient.
The KKT stationarity condition for robot $i$ is 
\begin{equation*}
    \nu u_i-\nu u_{0i}
    +\frac{1-\nu}{2}\sum_{j\neq i}E_{ij}
    -\lambda_{\mathrm{cbf}}
    [\mathcal{L}_{g_a}h_0]_i
    +(\boldsymbol{\mu}_{\mathrm{ub}})_i=0,
\end{equation*}
where $E_{ij}
:=
(q_{ij}+q_{ji})u_j
+(q_{ij}-q_{ji})u_m \geq 2q_{ij}u_m$, using $u_j\geq u_m$ and $q_{ij}+q_{ji}\geq0$.
Since $u_m<0<u_i$, robot $i$'s lower input constraint is inactive, and complementary slackness implies that its associated multiplier satisfies  $(\boldsymbol{\mu}_{\mathrm{lb}})_i=0$.
Moreover, since $\lambda_{\mathrm{cbf}}\geq0$  is the multiplier associated with the $g^{cbf}$ constraint, 
$[\mathcal{L}_{g_a}h_0]_i\leq0$, and $
(\boldsymbol{\mu}_{\mathrm{ub}})_i\geq0$ is the multiplier associated with $\mathbf{g}^{ub}$, the stationarity condition implies
$\nu u_i-\nu u_{0i}
+(1-\nu)u_m\sum_{j\neq i}q_{ij} \leq 0$. By the assumption on the nominal controller $u_{0i}(\boldsymbol{\xi}_i)$, this yields $u_i \leq 0$ whenever $v_i\geq V_i$. 
\end{proof}
\begin{rmk}
An exemplary nominal controller that satisfies condition of Proposition \eqref{compactness} is the proportional speed tracking controller $u_{0i}(\boldsymbol\xi_i)=K_p(v_{r}-v_i),$
where $v_r$ is the reference speed and $K_p>0$. 
Then, $V_i=v_{r} + \frac{(1-\nu)}{\nu K_p}|u_m|
\sum_{j\neq i}q_{ij}$, ensures that, for every $v_i\geq V_i$,
$u_{0i}(\boldsymbol\xi_i)
\leq \frac{(1-\nu)}{\nu}u_m
\sum_{j\ne i}q_{ij}$.
\end{rmk}
Next, by applying \cite[Theorem 3.1]{hager1979lipschitz}, 
the optimal solution of QP \eqref{QP_center_1} is shown to be locally Lipschitz continuous.
\begin{thm}\label{theorem2}
    In addition to conditions of Proposition \ref{compactness}, if $H \succ 0$, and the nominal controller $\boldsymbol{u}_0(\boldsymbol{\xi}_a)$ is locally Lipschitz continuous in $\boldsymbol{\xi}_a$, then the optimal solution $[\mathbf{u}_a(\boldsymbol{\xi}_a), \delta(\boldsymbol{\xi}_a)]^\top$ of QP \eqref{QP_center_1} is locally Lipschitz continuous. 
\end{thm}
\begin{proof}
Applying \cite[Theorem 3.1]{hager1979lipschitz} reduces to verifying assumptions \textbf{A1}–\textbf{A4} for the QP
\begin{equation}\label{Lip_QP}
     \min_{ \mathbf y}  \  \frac{1}{2} \mathbf y^\top R \mathbf y + r(\boldsymbol{\xi}_a)^\top \mathbf y, 
    \ \text{s.t.} \ A(\boldsymbol{\xi}_a) \mathbf y + b(\boldsymbol{\xi}_a) \le 0,
\end{equation}
where $\mathbf{y} = [\mathbf{u}_a^\top, \delta]^\top, R =diag(H,p), r(\boldsymbol{\xi}_a)=[l(\boldsymbol{\xi}_a), 0]^\top$, and the constraint matrix $A \in \mathbb{R}^{(2N+1) \times (N+1)}$ and vector $b \in \mathbb{R}^{2N+1}$ are
$
A(\boldsymbol{\xi}_a) =-
\begin{bmatrix} 
\mathcal{L}_{g_a} h_0 & 1 \\ 
I  & 0_{N\times 1}\\ 
-I & 0_{N\times 1}\\ 
\end{bmatrix}, 
b(\boldsymbol{\xi}_a) =
\begin{bmatrix} 
-\mathcal{L}_{f_a} h_0 - \gamma h_0 \\ 
u_m\mathbf{1}_N \\ 
-u_M\mathbf{1}_N \\ 
\end{bmatrix},
$
with $0_{N\times 1}$ being a zero column vector of length $N$.  Specifically, \textbf{A1}: for every $\boldsymbol{\xi}_a$, \eqref{Lip_QP} has a unique solution; \textbf{A2}: for every $\boldsymbol{\xi}_a$, there exists $\Gamma_1 <\infty$ such that $\|M\|_2 < \Gamma_1$, where $M$ is the active constraint matrix at $\boldsymbol{\xi}_a$;
\textbf{A3}: there exists $\omega>0$ such that for all $\boldsymbol{\xi}_a$, $\mathbf y^\top R \mathbf y \geq \omega \|\mathbf y\|_2^2$ for all $\mathbf y$ satisfying $M \mathbf y =0$; and \textbf{A4}: there exists $\mu>0$ such that $\|M^\top \zeta \|_2 \geq \mu \| \zeta\|_2$ for all $\zeta$ and $\boldsymbol{\xi}_a$.

Since $H \succ 0$ and $p > 0$, then $R = diag(H,p) \succ 0$, and hence \textbf{A1} is satisfied. 
% Moreover, $\|R\|_2 =\lambda_{\max}(R) = \max \{ \lambda_{\max}(H), \; p\}< \infty$, thereby \textbf{A2} is satisfied with $\Gamma_1 = \lambda_{\max}(R)$.
Since $M$ is formed by selecting rows of $A(\boldsymbol{\xi}_a)$, $\|M\|_2 \le  \|A(\boldsymbol{\xi}_a)\|_2$. 
The matrix $A(\boldsymbol{\xi}_a)$, and hence its spectral norm 
$\|A(\boldsymbol{\xi}_a)\|_2$, is continuous in $\boldsymbol{\xi}_a$. By compactness of the domain of $\boldsymbol{\xi}_a$, $\|A(\boldsymbol{\xi}_a)\|_2$ attains a finite maximum. Thus, \textbf{A2} holds with $\Gamma_1=\max_{\boldsymbol{\xi}_a}\|A(\boldsymbol{\xi}_a)\|_2<\infty$.

Since $R^\top = R\succ 0$, \textbf{A3} holds with $\omega = \lambda_{min}(R) >0$, where $\lambda_{min}(\cdot)$ denotes the minimum eigenvalue of a matrix. For \textbf{A4}, observe that $\|M^\top\zeta\|^2_2 = \zeta^\top M M^\top \zeta \ge \lambda_{\min}(M M^\top) \|\zeta\|^2_2$.
When all constraints are active, the active constraint matrix is $M_{all} = \begin{bmatrix}
\mathcal{L}_{g} h_0 & 1 \\ 
D_s & 0_{N\times 1}
\end{bmatrix}$, $D_s=\operatorname{diag}(\pm1,\ldots,\pm1)$, has linearly independent rows.
Since every active constraint matrix $M$ is a row submatrix of $M_{all}$, its rows are also linearly independent.
Therefore, $\lambda_{\min}(MM^T)>0$ for every $\boldsymbol{\xi}_a$. Since $\mathcal{L}_{g_a}h_0$ is continuous and the domain of $\boldsymbol{\xi}_a$ is compact, $\lambda_{\min}(MM^T)$ attains a strictly positive minimum over $\boldsymbol{\xi}_a$.
Since there are finitely many possible active sets, there exists a uniform $\mu>0$ such that $\lambda_{\min}(MM^T) \ge \mu^2$ for all $\boldsymbol{\xi}_a$ and all active constraint matrix $M$.
Since $h_0$ is twice continuously differentiable and $g_a$ is locally Lipschitz, $\mathcal{L}_{g_a}h_0$ is locally Lipschitz. Moreover, $r$ is locally Lipschitz in $\boldsymbol{\xi}_a$ due to the local Lipschitz continuity of the nominal controller $\mathbf{u}_0(\boldsymbol{\xi}_a)$. Therefore, by \cite[Theorem 3.1]{hager1979lipschitz}, for every $\boldsymbol{\xi}_{a}$, there exist a neighborhood $\mathcal{N}(\boldsymbol{\xi}_{a})$ and a Lipschitz constant $L_d < \infty$ such that, for all $\boldsymbol{\xi}_{a,1}, \boldsymbol{\xi}_{a,2} \in \mathcal{N}(\boldsymbol{\xi}_{a})$, 
$\| [\mathbf{u}^\top_a(\boldsymbol{\xi}_{a,1}), \delta(\boldsymbol{\xi}_{a,1})]^\top -[\mathbf{u}^\top_a(\boldsymbol{\xi}_{a,2}), \delta(\boldsymbol{\xi}_{a,2})]^\top\|_2 \le L_d\|\boldsymbol{\xi}_{a,1}-\boldsymbol{\xi}_{a,2}\|_2$. 
Thus, $\mathbf{y}$ is locally Lipschitz in $\boldsymbol{\xi}_{a}$.
\end{proof}
Note that restricting the slack variable to $\delta \le 0$ ensures forward invariance of the set  $\{\boldsymbol{\xi}_a \ | \ h_{0}(\boldsymbol{\xi}_a) \geq 0 \}$, but prevents direct application of \cite[Theorem 3.1]{hager1979lipschitz}. Indeed, when the CBF constraint, the constraint $\delta \le 0$, and $N$ input constraints are simultaneously active, $M$ has $N+2$ rows and $N+1$ columns. Thus, its rows are linearly dependent, violating \textbf{A4}. 

\noindent \textbf{Forward
invariance guarantees:}
The following result establishes a uniform bound on the optimal sign-indefinite slack $\delta(\boldsymbol{\xi}_a)$, characterizes the resulting forward-invariant set, and characterizes how the conservative margin $d_0$ should be selected to preserve a desired minimum safety margin $\tilde{d}$.
\begin{thm}\label{forwardinv}
    Under the conditions of Theorem \ref{theorem2}, there exists a finite $\delta_m \geq 0$ such that $\delta(\boldsymbol{\xi}_a) \leq \delta_m$
    uniformly over the compact domain $\mathcal{X}$. If, in addition, $d_0^2 \geq  \tilde d^2 +  \frac{\delta_m}{\gamma}$, then the compensated set
    $\mathcal C_{d_0}: = \{ \boldsymbol{\xi}_a \ | \ h_0(\boldsymbol{\xi}_a) +\frac{\delta_m}{\gamma}\geq0\}$ is forward invariant and $\mathcal C_{d_0} \subset \tilde {\mathcal{C}}$.
\end{thm}
\begin{proof}
By Theorem \ref{theorem2}, the slack
$\delta(\boldsymbol{\xi}_a)$ is continuous. Therefore, on the compact
operating domain $\mathcal X$, it admits the finite uniform bound
$\delta_{m}$. Next, for each $i \in [N]$, observe that $h(\boldsymbol{\xi}_i; d_0) \leq h(\boldsymbol{\xi}_i; \tilde d)-\frac{\delta_m}{\gamma}$. 
Consequently, 
$h_{ij}(\boldsymbol{\xi}_a;\kappa, d_0) \le h_{ij}(\boldsymbol{\xi}_a;\kappa, \tilde d)- \frac{\delta_m}{\gamma}$ for all $j \in [N], \ j\ne i$.
Hence, it follows that
$h_0(\boldsymbol{\xi}_a) + \frac{\delta_m}{\gamma}\leq \tilde h(\boldsymbol{\xi}_a)$, which implies 
$\mathcal C_{d_0} \subset \tilde {\mathcal C}$.
Finally, the $g^{cbf}$ constraint imposed by the QP \eqref{QP_center_1} yields $\frac{d}{dt}h_0 + \gamma h_0 \geq -\delta(\boldsymbol{\xi}_a) \geq -\delta_{m}$. Integrating over $[0,t]$ yields $h_0(\boldsymbol{\xi}_a(t)) + \frac{\delta_{m}}{\gamma} \ge e^{-\gamma t} \big(h_0(\boldsymbol{\xi}_a(0)) + \frac{\delta_{m}}{\gamma} \big) $. Thus, if $\boldsymbol{\xi}_a(0) \in \mathcal{C}_{d_0}$, then  $\boldsymbol{\xi}_a(t) \in \mathcal{C}_{d_0}, \forall t\geq 0$. 
\end{proof}
The following remark highlights the tradeoff of introducing a sign-indefinite slack: increasing $p$ cannot eliminate the positive slack due to finite maximum deceleration, leading to a more conservative forward-invariant set $\mathcal C_{d_0}$.

\begin{rmk}
To illustrate why $\delta >0$ may be unavoidable, consider a state
satisfying $-\frac{\delta_m}{\gamma}
\leq h_0(\boldsymbol{\xi}_a)<0$ and $\mathbf v_a=\mathbf0$, for which $\mathcal{L}_{g_a}h_0= \mathbf0$ and $\mathcal{L}_{f_a}h_0=0$. Hence,
$\gamma h_0(\boldsymbol{\xi}_a)<0$, requiring $\delta>0$ to satisfy the $g^{cbf} \geq 0$. 
Moreover, since $\mathcal{L}_{g_a}h_0\preceq 0$, then $\max\limits_{\mathbf{u}_a \in [u_m, u_M]^N} g^{cbf} = F + \gamma h_0 + \delta$, where $F:= \mathcal{L}_{f_a}h_0 + \mathcal{L}_{g_a} h_0^\top u_m\mathbf{1}_N$. For $\boldsymbol{\xi}_a(0) \in \mathcal{C}_{d_0}$, the minimum nonnegative slack required to satisfy $g^{cbf} \geq 0$ is $\delta_{\mathrm{r}}(\boldsymbol\xi_a)
:= \left[-F(\boldsymbol\xi_a)-\gamma h_0(\boldsymbol\xi_a)\right]_+$,  which is independent of $Q$ or $p$. Over the compact
set $\mathcal X \cap\mathcal{C}_{d_0}$, define
$\delta_{\mathrm{rm}}
:=\max_{\boldsymbol\xi_a}\delta_{\mathrm{r}}(\boldsymbol\xi_a),$ and 
$B:=\max_{\boldsymbol\xi_a}
\left(\eta(u_m\mathbf1_N,\boldsymbol\xi_a)
-\underline m(\boldsymbol\xi_a)\right),$
where $\underline m(\boldsymbol\xi_a):=-\frac12 l^\top H^{-1} l$, and $\eta(\mathbf{u}_a,\boldsymbol\xi_a):= \frac{1}{2}\mathbf{u}_a^\top H\mathbf{u}_a + l^\top\mathbf{u}_a$.
Since $(u_m\mathbf1_N,\delta_{\mathrm{r}}(\boldsymbol \xi_a))$ is feasible in the QP \eqref{QP_center_1} and $H\succ0$, optimality yields
$\delta_{\mathrm{r}}(\boldsymbol \xi_a)
\leq \delta_m \le 
\sqrt{
(\delta_{\mathrm{rm}})^2+
\frac{2B}{p}}.$
Thus, increasing $p$ cannot eliminate the slack $\delta_{\mathrm{rm}}$.
\end{rmk}

% \end{rmk}
\noindent \textbf{Deadlock-free operation:}
Next, sufficient conditions on $\mathbf{u}_a(\boldsymbol{\xi}_a)$ are established to guarantee $\mathcal{J}_D$ is empty.
\begin{Prop}\label{deadlock}
    If $\mathbf{u}_0 (\boldsymbol{\xi}_a) \succ 0$ for all $\boldsymbol{\xi}_a$ satisfying $\mathbf{v}_a = \mathbf{0}$, and there exists at least one $k$ such that $q_{kj} > q_{jk}$, $q_{kj}>0 \; \forall j \neq k$, then $\mathcal{J}_D = \varnothing$.
    % then the optimal solution of \eqref{QP_center_1} satisfies $u_i^*(\boldsymbol{\xi}_a) > 0$. 
\end{Prop}
\begin{proof}
Assume, by contradiction, that $\mathcal{J}_D\neq\varnothing$ and let  $\boldsymbol{\xi}_a\in\mathcal{J}_D$. If $u_i(\boldsymbol{\xi}_a)=u_M>0$ for some robot $i$, then $\tilde{\mathbf{u}}(\boldsymbol{\xi}_a)\neq\mathbf{0}$, yielding to a contradiction. It remains to consider the case in which
$u_i(\boldsymbol{\xi}_a)\neq u_M,\forall i \in [N]$.
Note that $\mathcal{L}_{g_a} h_0(\boldsymbol{\xi}_a)=0$, for $\boldsymbol{\xi}_a\in\mathcal{J}_D$. Hence, the KKT stationarity condition simplifies to $\big(\nu I  + (1-\nu)Q_{sp} \big)\mathbf{u}_a + l - \boldsymbol{\mu}_{\mathrm{lb}}=0$, where $\boldsymbol{\mu}_{\mathrm{lb}} \geq 0$ and the multiplier associated with $\boldsymbol{g}^{ub}$ constraint is zero because $u_i(\boldsymbol{\xi}_a)\neq u_M,\forall i \in [N]$. It remains to show that $u_k(\boldsymbol{\xi}_a) > 0$, for $\boldsymbol{\xi}_a\in\mathcal{J}_D$, from which the conclusion $\mathcal{J}_D = \varnothing$ follows immediately. 
For robot $k$, the KKT stationary condition reduces to $\nu u_k + \frac{1-\nu}{2}\sum\limits_{j\neq k} E_{kj} - (\boldsymbol{\mu}_{\mathrm{lb}})_k= \nu u_{0k} >0$, where $E_{kj} = (q_{kj}+ q_{jk})u_j + (q_{kj} - q_{jk})u_m$. It suffices to consider $u_j(\boldsymbol{\xi}_a) \in [u_m, 0], \;  \forall j \neq k$, since the conclusion $\mathcal{J}_D = \varnothing$  follows immediate otherwise.
Under the assumptions $q_{kj} > q_{jk}$ and $q_{kj}>0$, for $\forall j \neq k$, we have $E_{kj}<0$. 
% Indeed, if $q_{ij}+ q_{ji} \geq 0$, $\sup_{u_j \in [u_m, 0]} E_{ij} = (q_{ij} - q_{ji})u_m <0$, whereas if $q_{ij}+ q_{ji} < 0$, $\sup_{u_j \in [u_m, 0]} E_{ij} = 2q_{ij}u_m<0$. 
Consequently, $u_k(\boldsymbol{\xi}_a) > 0$, for  $\boldsymbol{\xi}_a\in \mathcal{J}_D$. 
% , implying that $\mathcal{J}_D = \varnothing$. 
    % Next, suppose $\mathbf{u}_a = u_m\mathbf{1}_2$. Then $H \mathbf{1}_2u_m - \nu \mathbf{u}_0 - \boldsymbol{\mu}_{lb}=0$.
\end{proof}
% for a given $p$, $\gamma$ and $\u_b$ by calculating $\delta_m$ offline for a given p, gamma and u_b . 
The control parameters can be selected jointly as follows. First, choose
$Q$ such that conditions in Propositions~\ref{prop2},~\ref{compactness}, and~\ref{deadlock} hold.
% To guarantee deadlock-free operation, require at least one robot
% $k\in[N]$ to satisfy $q_{kj}>q_{jk},  q_{kj}>0, \forall j\neq k.$
Then, for candidate control parameters of $(p,\gamma, u_b)$, and
$d_0>\tilde d$, compute
$\delta_m$ offline and
retain the parameters satisfying
$d_0^2-\tilde d^2 \geq \frac{\delta_m}{\gamma}.$

%% file: Simulation.tex
\section{SIMULATION RESULTS}\label{Simulation}
This section presents the simulation results for two robots navigating a conflict zone, as shown in Fig. \ref{Intro_collision}. Each robot uses an onboard nominal controller $u_{0i}(\boldsymbol{\xi}_i)=K_p(v_r-v_i)$, where $v_r= 3[m/s]$ is a common reference speed and $K_p=1.2$.
The simulation parameters are set as follows: $d=1[m]$, $d_0=4[m]$, $[u_m,u_M] = [-3.4,3]$[$m/s^2$], $u_b=0.3u_m$, $\gamma=1.3$, $v_{\mathrm{th}}=0.1$[$m/s$], and $p=1000$.
Simulations were performed on a MacBook Pro (M1, 16 GB RAM), with QP \eqref{QP_center_1} solved using CVXOPT solver in Python that yields an average computation time of $1.4[ms]$. 
\begin{figure}[t]
    \centering
    % Row 1
    \begin{subfigure}[t]{0.489\columnwidth}
    \centering
        \includegraphics[width=\linewidth]{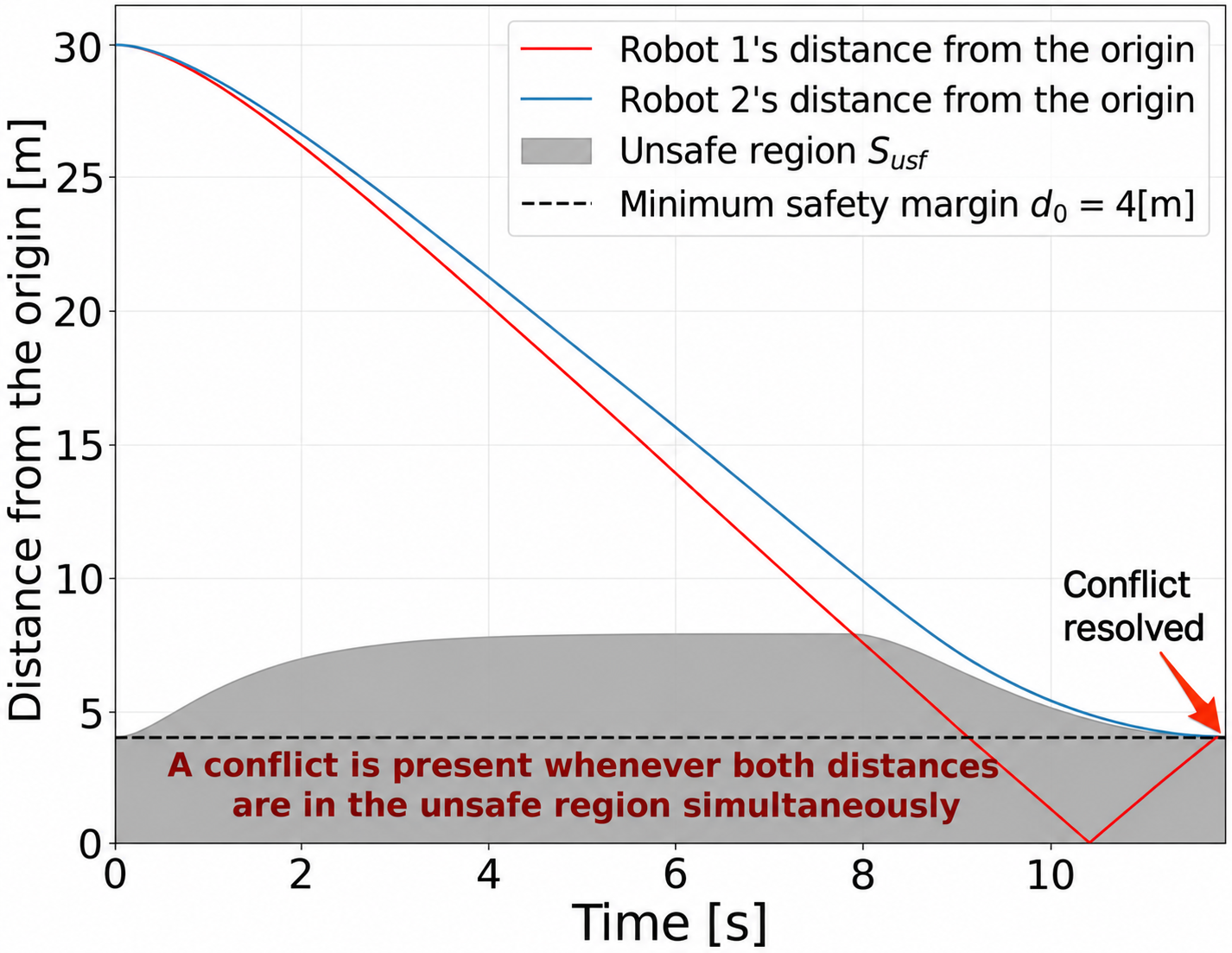}
        \caption{}
    \end{subfigure}
     \hfill
       \begin{subfigure}[t]{0.48\columnwidth}
       \centering
        \includegraphics[width=\linewidth]{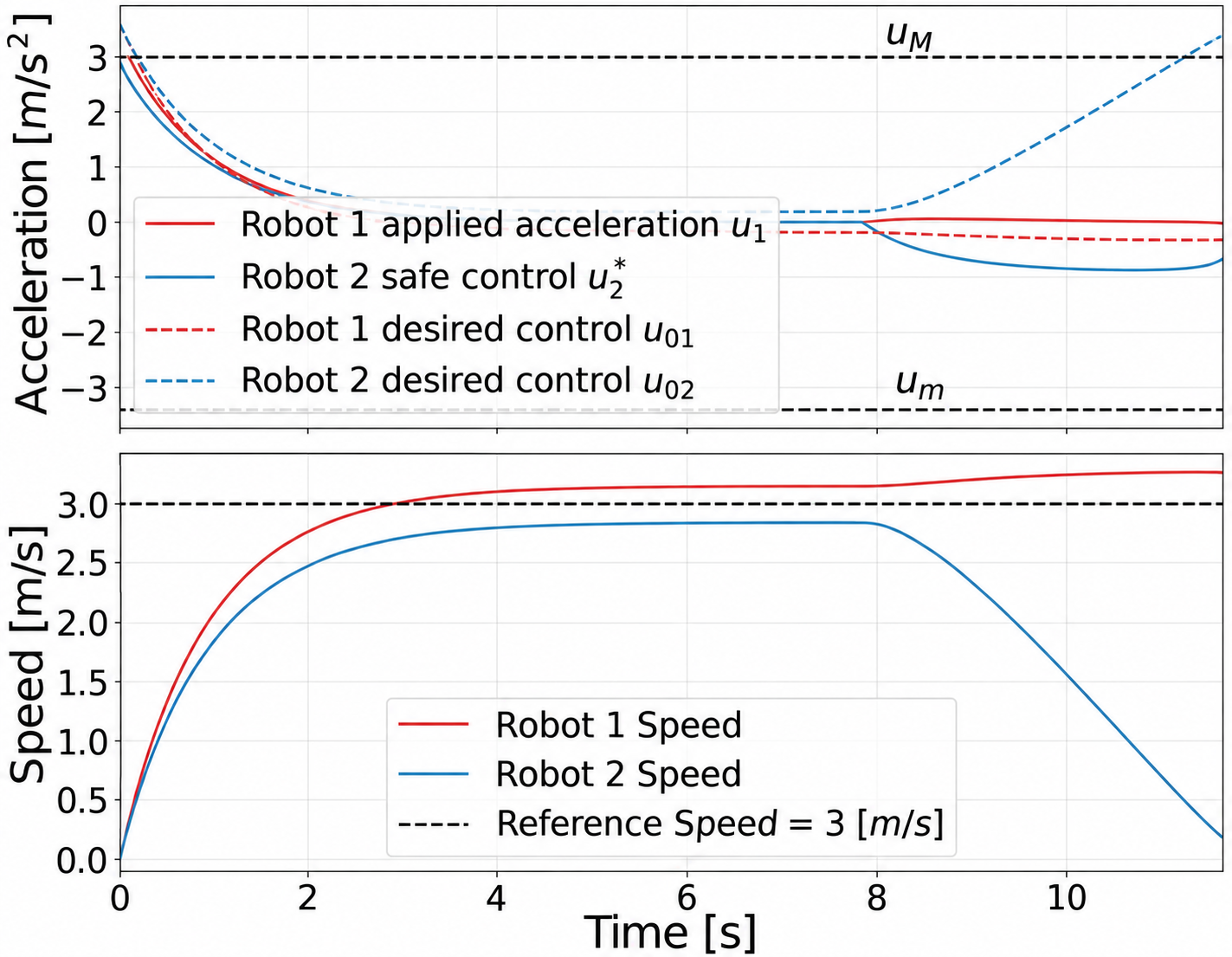}
        \caption{}
    \end{subfigure}
% \vspace{0.3em} % small vertical spacing between rows
\begin{subfigure}[t]{0.50\columnwidth}
\centering
        \includegraphics[width=\linewidth]{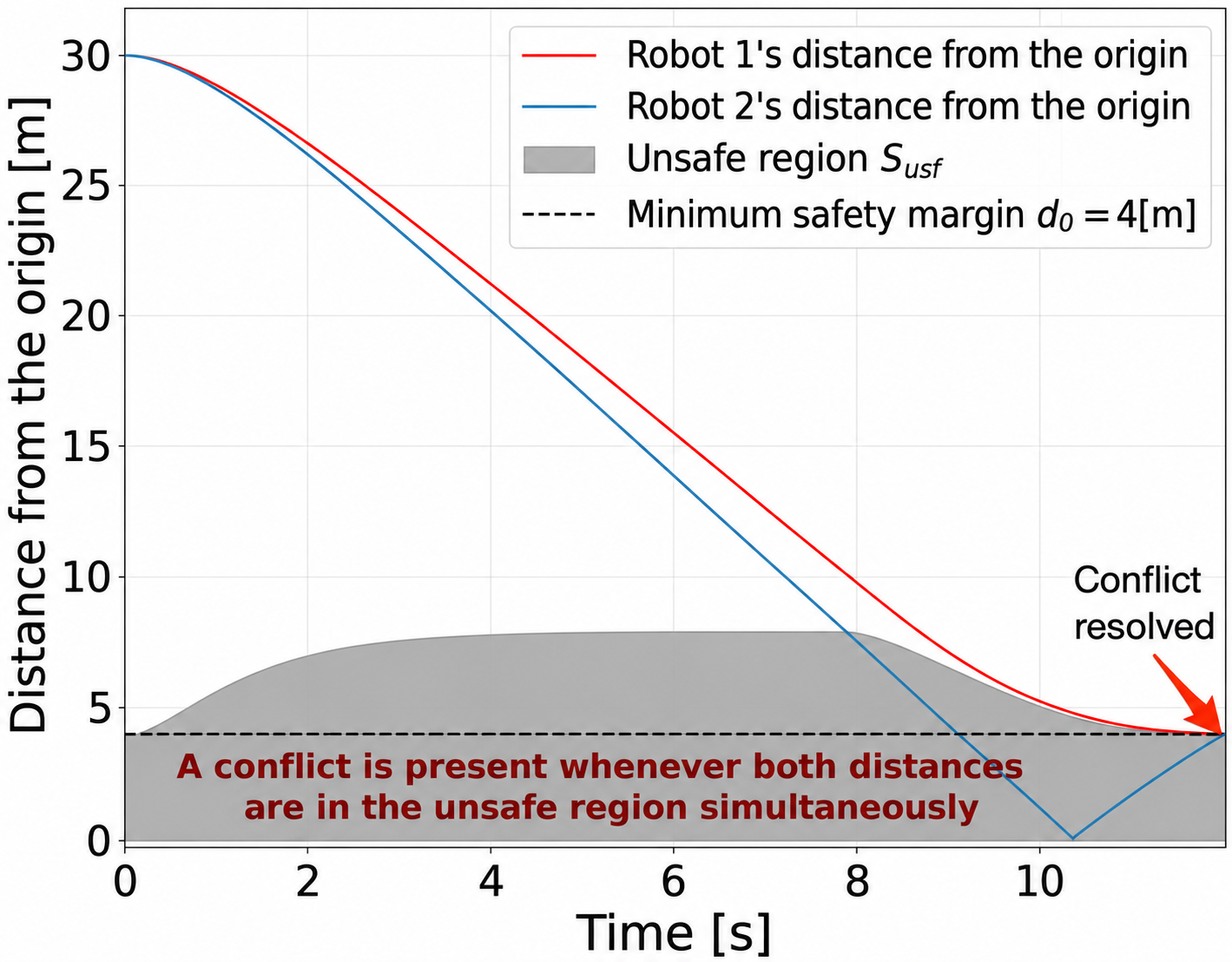}
        \caption{}
    \end{subfigure}
     \hfill
    \begin{subfigure}[t]{0.47\columnwidth}
    \centering
        \includegraphics[width=\linewidth]{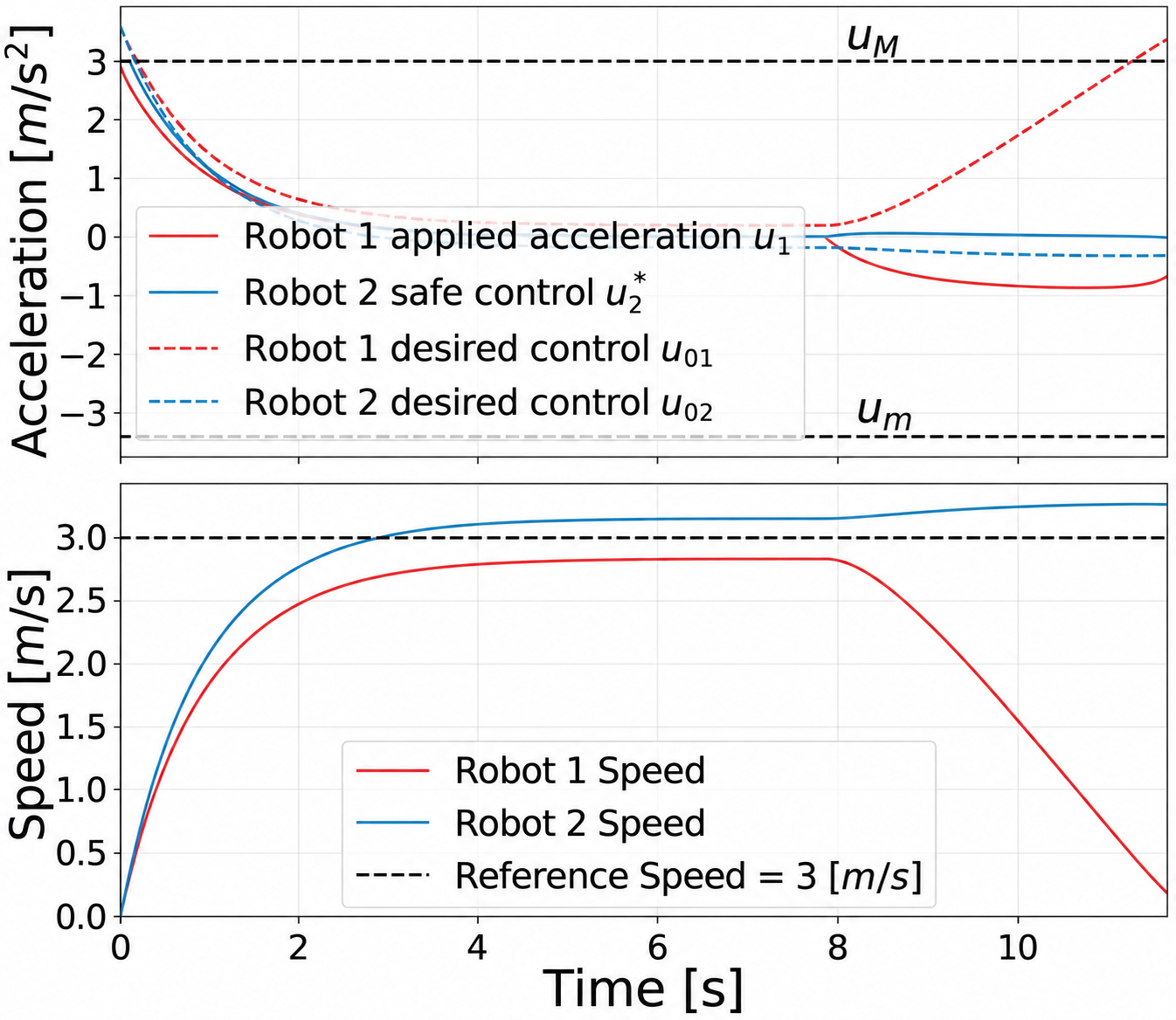}
         \caption{}
    \end{subfigure}
   \caption{(a)-(b):
   $\nu = 0.9$, robots 1 and 2 yield preferences: $q_{12} = 2$, $q_{21} = 1$. Robot distances from the conflict-zone center are shown. The red arrow shows robot $1$ exiting a circle of radius $d_0$ centered at the origin of the conflict zone, after which the conflict with robot $2$ is resolved. (c)-(d): $\nu = 0.9$, with swapped yield preferences $q_{12} = 1$, $q_{21} = 2$, showing robots’ speed trajectories for $v_r = 3[m/s]$.}
    \label{fig:result1}
    \vspace{-0.2cm}
\end{figure}
Fig. \ref{fig:result1}(a)–(d) show the robots’ motion under two yield preferences with $\nu =0.9$, favoring the nominal controller that drives both robots toward the common reference speed $v_r$. Both robots start symmetrically $30[m]$ from the origin at rest. In Figs. \ref{fig:result1}(a)–(b), robot 1 has higher yield
preference ($q_{12}=2$) than robot 2 ($q_{21}=1$), whereas in \ref{fig:result1}(c)–(d) the yield preferences are reversed ($q_{12}=1$, $q_{21}=2$), resulting in a reversed traversal order through the unsafe region.
As shown in Figs. \ref{fig:result1}(a)–(b), for both cases, no collision occurs, as the robots never simultaneously enter the unsafe region shown in gray. 
With $\nu =0.9$ favoring nominal control inputs, both robots initially accelerate toward $v_r$, as shown in Figs. \ref{fig:result1}(b) and (d), then one yields near the unsafe region by slowing down, as seen in the speed profiles.
Fig. \ref{fig:combined2}(a) shows the robots’ motion for $\nu=0.1$, prioritizing the coupling cost over nominal speed tracking. Robot 2 ($q_{21} =0.05$) yields to robot 1 ($q_{12} =0.1$), and both start symmetrically $30[m]$ from the origin at rest. 
Compared to Fig. \ref{fig:result1}(a), larger coupling cost prioritization leads to a larger separation in conflict zone crossing times, causing robot 2 to enter the conflict zone later and increasing its travel time.
Fig.~\ref{fig:combined2}(b) shows a large initial difference between the control inputs caused by robot~2’s yielding response, which decreases once persistent speed separation is established.
In contrast, in Fig. \ref{fig:result1}(b), with $\nu = 0.9$ favoring nominal speed tracking, both robots initially accelerate toward the reference speed; near the unsafe region, robot 2 yields by slowing down, allowing robot 1 to pass first.
% Average total travel times
\begin{figure}
    \centering
    % First figure
    \begin{subfigure}{0.5\columnwidth}  % Reduced width to fit 4 in a row
        \includegraphics[width=\linewidth]{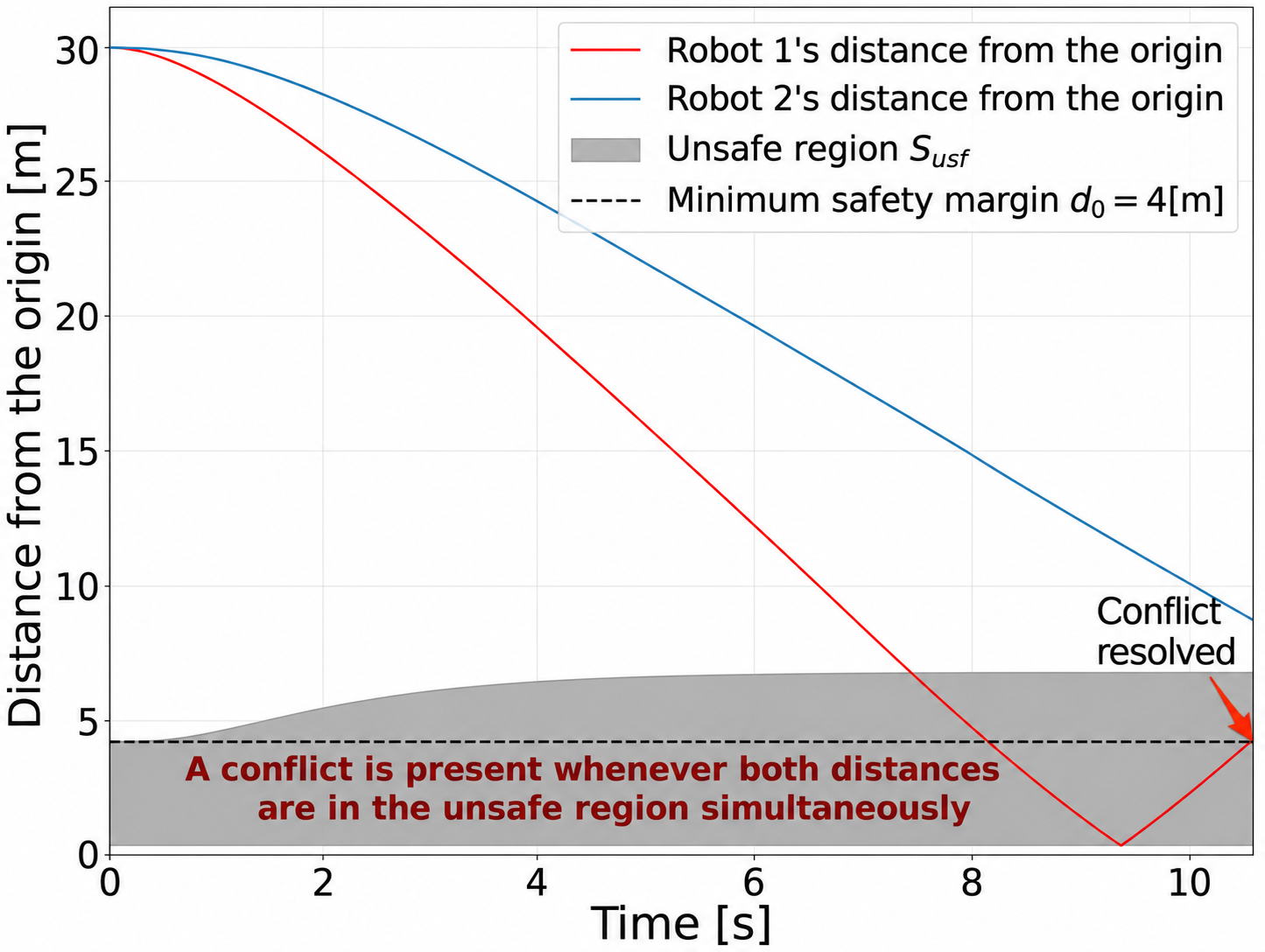}
        \caption{}
    \end{subfigure}
    % Second figure
    % \begin{subfigure}{0.49\columnwidth}  % Reduced width to fit 4 in a row
    %     \includegraphics[width=\linewidth]{fig/Distance_from_center_4.png}
    %     \caption{Case 2}
    % \end{subfigure}
    % Third figure
    \begin{subfigure}{0.478\columnwidth}  % Reduced width to fit 4 in a row
        \includegraphics[width=\linewidth]{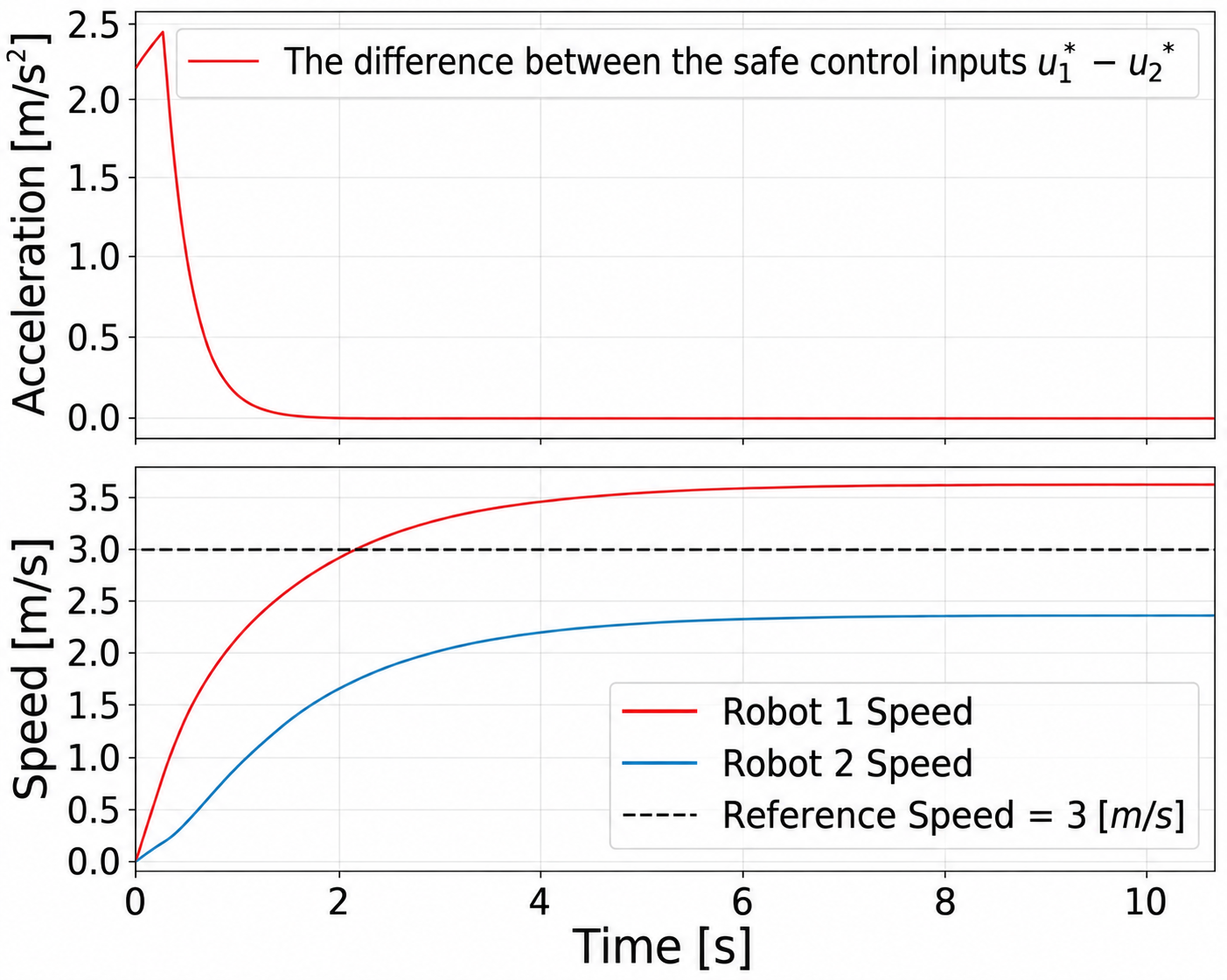}
        \caption{}
    \end{subfigure}
    % Fourth figure
    % \begin{subfigure}{0.47\columnwidth}  % Reduced width to fit 4 in a row
    %     \includegraphics[width=\linewidth]{fig/Speed_acc_4.png}
    %     \caption{Case 4}
    % \end{subfigure}
    \caption{(a)-(b): Robots 1 and 2 yield
    preferences: $q_{12} = 0.1$, $q_{21} = 0.05$. (a) For $\nu = 0.1$, the exit time of yielding robot 2 increases compared to Fig. \ref{fig:result1}, as yielding reduces its acceleration and, thus speed.
    (b) $\nu = 0.1$ provides greater weight on asymmetric coupling cost, producing a large initial acceleration difference.
    % CAV positions and control profiles under the proposed IC. The gray region marks a $R=2[m]$ radius around the conflict point.
    }
    \label{fig:combined2}
    \vspace{-0.1cm}
\end{figure}
% \begin{figure}
%     \centering
%     \includegraphics[width=2.7in]{fig/Total_transfer_times2.png}
%     \caption{\small Average minimum and total travel times denote the exit times of the first and last robots, respectively, from a circle of radius $d_0=4$[m], centered at the conflict zone origin, averaged over initial conditions sampled from the safe set $\mathcal{C}$ with trade-off parameter $\nu =0.9$.}
%     \label{Fig:TransferTime}
% \end{figure}

\begin{figure}[t]
    \centering
    \begin{minipage}[c]{0.52\columnwidth}
        \centering
        \includegraphics[width=\linewidth]{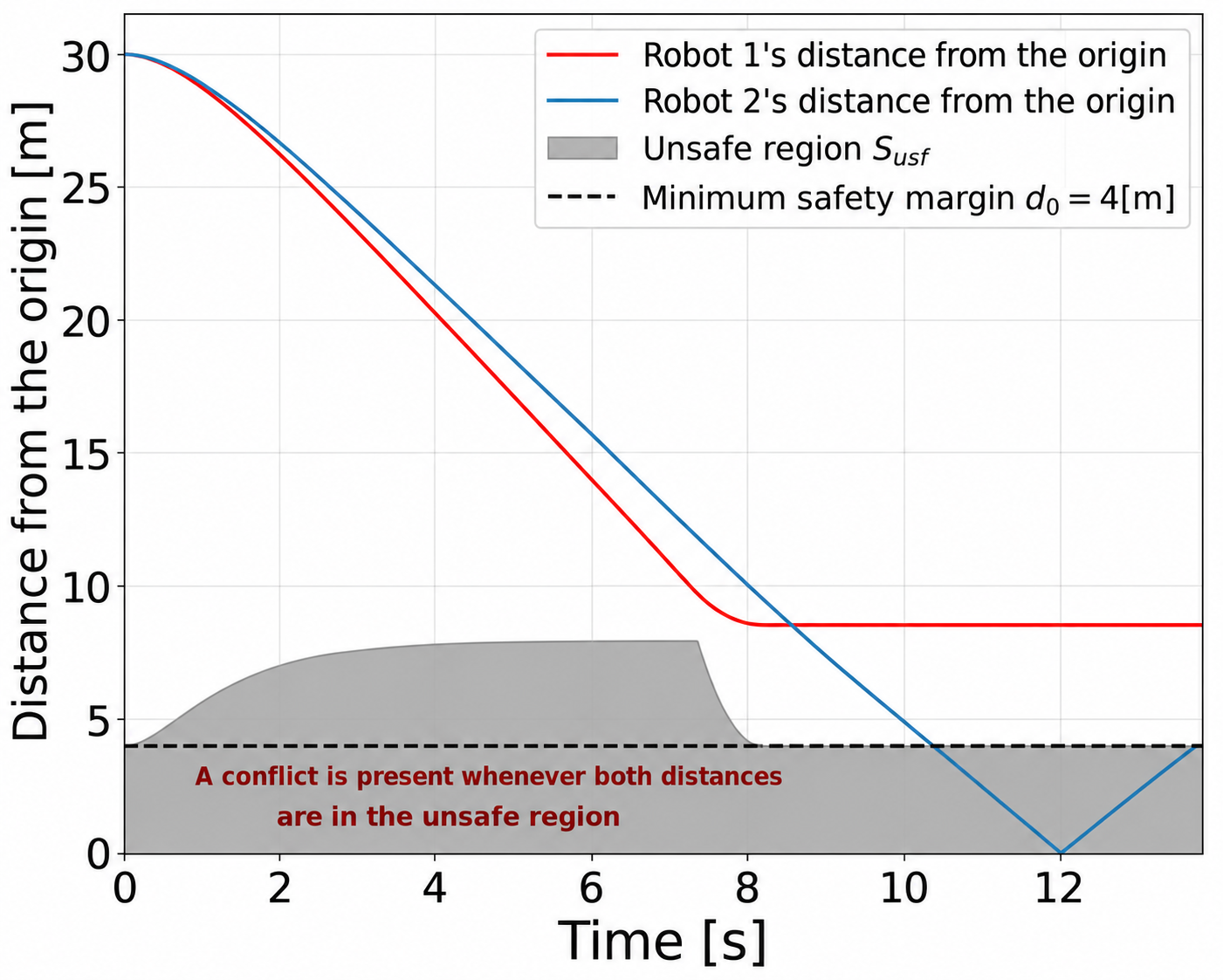}
    \end{minipage}
    \hfill
    \begin{minipage}[t]{0.40\columnwidth}
        \vspace{-1.8cm}
        \caption{$\nu = 0.9$, with robots 1 and 2 yield
        preferences: $q_{12} = 2$, $q_{21} = 1$, inducing robot 2 to yield.
        When robot 1 fails at $10[m]$ from the origin, it initiates emergency braking $u_m = -3.4[m/s^2]$.}
        \label{Fig:mulfunction}
    \end{minipage}
     \vspace{-0.4cm}
\end{figure}

Fig. \ref{Fig:mulfunction} shows a scenario where robot 1, with higher priority ($q_{12} = 2$), fails at $10[m]$ from the origin and no longer follows the commanded acceleration $u_1(\boldsymbol{\xi}_a)$. The failure is represented as emergency braking triggered by a fault, such as an obstacle detection upon which robot 1 applies maximum braking $u_m$. 
Fig. \ref{Fig:mulfunction} shows robust coordination under robot failure: despite robot 2 initially yielding to robot 1, it proceeds first after robot 1 fails, reflecting the adaptive coordination enabled by the yielding matrix. If robot 1 fails within the unsafe region, robot 2 stops at the boundary of the unsafe region, thereby preserving safety.
% The intersection shown in Fig. 1 was built with $S_c = 20 [m]$ and $S_o = 40 [m]$, lane width of $3[m]$, and the speed limit was set to $vcs = 15 [m/s] ≈ 33.5 [mph]$
% \subsection{Simulation results of centralized coordination}
% \subsection{Simulation results of decentralized coordination}

%% file: Conclusion.tex
\section{CONCLUSION}\label{CONCLUSION}
This paper presents a QP-based feedback controller for multi-robot conflict resolution with yield preferences. The controller integrates safety and yielding dependent coordination within a unified formulation while respecting input constraints. Sufficient conditions ensuring deadlock-free operation and local Lipschitz continuity of the controller are provided.
Simulations demonstrate coordination across diverse yielding preferences and robustness to robot failures. 
% Future work will consider state-dependent priorities and extensions to networks of coupled conflict zones, including intersections, junctions, and roundabouts.
% A unified negotiation and control framework for multi-robot conflict resolution is proposed using coupled CBF to ensure safety via forward invariance. 
% Collisions are avoided by minimally adjusting nominal controls through an input constrained QP, whose solution is locally Lipschitz and balances individual objectives with cooperation.
% Incorporating a cooperation matrix enables inter-robot coupling and introduces \lq soft\rq\ priority assignments for flexible coordination. Sufficient conditions for guaranteed conflict resolution are derived, and simulations demonstrate adaptability under varying priorities, yielding symmetric coordination when balanced and controlled asymmetry when priorities differ, highlighting its effectiveness in diverse multi-robot interaction scenarios.
% Future work will extend the approach to state-dependent priorities to improve total travel times and to multiple coupled conflict zones, including intersections, T- and Y-junctions, roundabouts, and their networks.